\documentclass[10pt,twocolumn,letterpaper]{article}

\usepackage[pagenumbers]{cvpr} 

\usepackage{graphicx}
\usepackage{amsmath}
\usepackage{amssymb}
\usepackage{booktabs}
\usepackage{multirow}

\usepackage[pagebackref,breaklinks,colorlinks]{hyperref}
\usepackage{xcolor}
\usepackage{siunitx}
\usepackage{enumitem}

\usepackage[capitalize]{cleveref}
\crefname{section}{Sec.}{Secs.}
\Crefname{section}{Section}{Sections}
\Crefname{table}{Table}{Tables}
\crefname{table}{Tab.}{Tabs.}
\usepackage{textpos}

\let\oldparagraph\paragraph
\renewcommand{\paragraph}[1]{\vspace{-0.2cm} \oldparagraph{#1}}
\newcommand{\figref}[1]{\mbox{Fig.~\ref{#1}}}
\newcommand{\tblref}[1]{\mbox{Table~\ref{#1}}}
\newcommand{\secref}[1]{\mbox{Sec.~\ref{#1}}}
\renewcommand{\eqref}[1]{\mbox{Eq.~\ref{#1}}}

\newcommand{\std}[1]{\scalebox{0.85}{$_{{\pm#1}}$}}

\begin{document}
\title{
Zero Experience Required: Plug \& Play Modular Transfer Learning \\for Semantic Visual Navigation
}

\author{Ziad Al-Halah$^{1}$ \hspace{5mm} Santhosh K. Ramakrishnan$^{1,2}$ \hspace{5mm} Kristen Grauman$^{1,2}$\\
$^1$The University of Texas at Austin \hspace{3mm} $^2$Meta AI
\\
{\footnotesize \texttt{ziadlhlh@gmail.com, srama@cs.utexas.edu, grauman@cs.utexas.edu}} \\
}
\maketitle

\begin{abstract}
In reinforcement learning for visual navigation, it is common to develop a model for each new task, and train that model from scratch with task-specific interactions in 3D environments.
However, this process is expensive; massive amounts of interactions are needed for the model to generalize well.
Moreover, this process is repeated whenever there is a change in the task type or the goal modality.
We present a unified approach to visual navigation using a novel modular transfer learning model.
Our model can effectively leverage its experience from one source task and apply it to multiple target tasks (\eg, ObjectNav, RoomNav, ViewNav) with various goal modalities (\eg, image, sketch, audio, label).
Furthermore, our model enables zero-shot experience learning, whereby it can solve the target tasks without receiving any task-specific interactive training.
Our experiments on multiple photorealistic datasets and challenging tasks show that our approach learns faster, generalizes better, and outperforms SoTA models by a significant margin. Project page: \url{https://vision.cs.utexas.edu/projects/zsel/}
\end{abstract}
 
\begin{textblock*}{\textwidth}(0cm,-16cm)
\centering 
IEEE Conference on Computer Vision and Pattern Recognition (CVPR), 2022.
\end{textblock*}

\section{Introduction}\label{sec:intro}

In visual navigation, an agent must intelligently move around in an unfamiliar environment
to reach a goal, using its egocentric camera to avoid obstacles and decide where to go next.
As a fundamental research problem in embodied AI, visual navigation has many potential applications---such as service robots in the home or workplace, mobile search and rescue robots, assistive technology for the visually impaired, and augmented reality systems to help people navigate or find objects.

\begin{figure}[t]
\centering
    \includegraphics[width=0.9\linewidth]{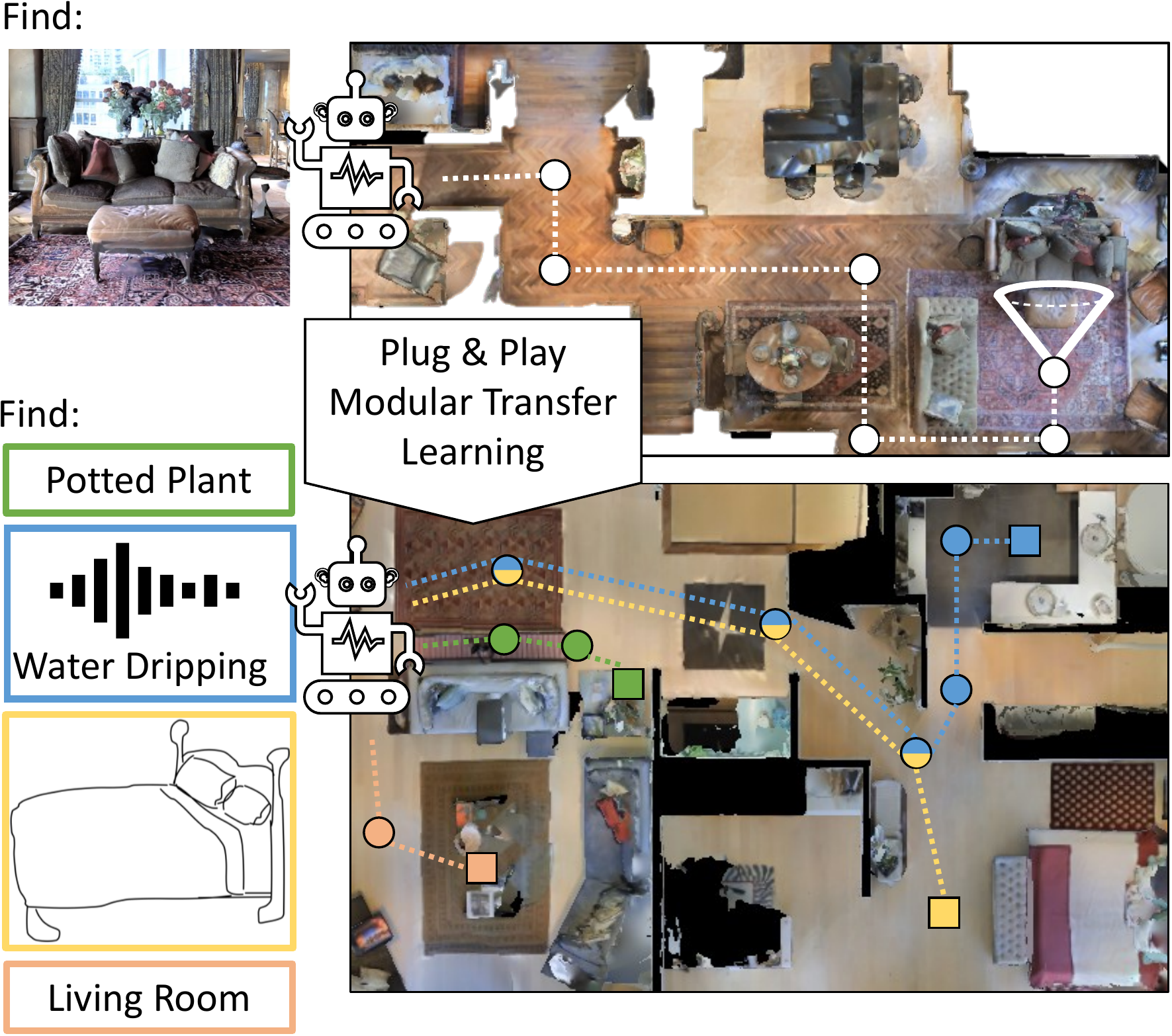}
\caption{
    Our novel modular transfer learning approach for semantic visual navigation learns a general purpose semantic search policy by finding image views sampled randomly in the environment (top).
    Then, this experience is leveraged to search for previously unseen types of goals and search tasks (bottom).
    Our approach enables zero-shot experience learning (\ie, perform the target task without receiving any new experiences) and it adapts its policy much faster using fewer target-specific interactions.
}
\label{fig:intro}
\vspace{-0.5cm}
\end{figure}
 
Recent work in computer vision explores visual navigation from many different fronts.  
In PointNav, an agent is asked to go to a specific position in an unmapped environment (\eg, go to $(x,y)$)~\cite{anderson2018evaluation,savva2019habitat,savva2017minos}.  
In ObjectNav, the agent must find an object by name (\eg, go to the nearest telephone)~\cite{batra2020objectnav,anderson2018evaluation}.  
In RoomNav, the agent must find a room (\eg, go to the kitchen)~\cite{wu2019,narasimhan2020roomnav,savva2017minos}.  
In AudioNav, the agent must find a sounding target (\eg, find the ringing phone)~\cite{chen2020audionav,gan2019}.  
In ImageNav, the agent must go to where a given photo was taken~\cite{zhu2017,chaplot2020neural,savinov2018semi}.  
Each case presents a distinct goal to the agent.  
Accordingly, researchers have pursued \emph{task-specific} models to treat each one, typically training policies with deep reinforcement learning (RL).

Despite exciting advances, learning task-specific navigation policies has inherent limitations.  
Training embodied agents from scratch for each new task and relying on special-purpose architectures and priors (\eg, room layout maps for RoomNav, object co-occurrence priors for ObjectNav, directional cues for AudioNav, etc.) requires repeated access to training environments for gathering new agent experience in the context of each task, greatly hindering sample efficiency.  
Even with today's fast simulators and photorealistic scanned environments~\cite{savva2019habitat,Matterport3D,xia2018gibson}, this typically amounts to days and weeks of computation on a small army of GPU servers to train a single policy.  
Moreover, by tackling each variant in isolation, agents fail to capture what is common across the tasks.  
Finally, some tasks require manual annotations such as object labels in 3D space, which naturally limits how extensively they can be trained.

In this work, we challenge the assumption that distinct navigation tasks require distinct policies.
Intuitively, finding a good policy for one navigation task should help with the rest.  
For example, if we know how to find a microwave, then finding a kitchen should be easy too; if we know how to find an object by name, then finding it based on a hand-drawn sketch---or the sounds it emits---should be possible too.  
In short, it should be beneficial to learn one navigation task and then apply the accumulated experience to many.

To that end, we propose a modular transfer learning approach for semantic visual navigation that enables \emph{zero-shot experience learning}.  See Figure~\ref{fig:intro}.
First, we develop a general-purpose semantic search policy.
Specifically, using a novel reward and task augmentation strategy, we train a source policy for the \emph{image-goal} task, where the agent receives a picture taken at some unknown camera pose somewhere in the environment, and must travel to find it.  
Next, we develop a joint goal embedding that is trained offline (\ie, no interactive agent experience) to relate various target goal types to image-goals.
Finally, we address target downstream tasks either by zero-shot transfer with no new agent experience, or by fine-tuning with a limited amount of agent experience on the target task.  

Zero-shot learning traditionally focuses on supervised tasks such as image recognition~\cite{lampert2009,al-halah2016,xian2017}, where models forgo using labeled samples for the new class.  
Instead, the proposed zero-shot \emph{experience} learning (ZSEL) focuses on reinforcement learning tasks, where models forgo using interactions in the physical environment for the new navigation task.  
ZSEL is important for lifelong learning, where an agent will face novel tasks once it is deployed and must solve them while using no or few training episodes.

Using hundreds of multi-room environments from Matterport3D~\cite{Matterport3D},  Gibson~\cite{xia2018gibson}, and HM3D~\cite{ramakrishnan2021hm3d}, we demonstrate our approach for four challenging tasks and goals expressed with five different modalities---images, category names, audio, hand-drawn sketches, and edgemaps.  
Our ImageNav results advance the state of the art, and our modular transfer approach outperforms the best existing methods of transfer based on self-supervision, supervision, and RL.
Finally, our ZSEL performance on $5$ semantic navigation tasks is equivalent to $507$ million interactions required by task-specific policies learned from scratch.

\section{Related Work}\label{sec:related}
\vspace{0.2cm}

\paragraph{Visual Navigation}
Traditional methods in visual navigation often rely on mapping the 3D space and then planning their movements~\cite{bailey2006,thrun2002,fuentes2015}.
However, fueled by fast simulators~\cite{savva2019habitat,kolve2017thor} and large-scale photorealistic  datasets~\cite{xia2018gibson,Matterport3D,chen2020audionav,ramakrishnan2021hm3d} there have been great advances in learning-based navigation approaches~\cite{chen2021avwan,chaplot2020neural,ramakrishnan2020occant} leading to a near-perfect agent for tasks like point-goal navigation~\cite{wijmans2020ddppo}.
In this work, we consider semantic visual navigation, where the agent is given a semantic description of the goal (\eg, object-goal~\cite{anderson2018evaluation,batra2020objectnav,ramakrishnan2022poni}, image-goal~\cite{zhu2017,chaplot2020neural}, room-goal~\cite{wu2019}, audio-goal~\cite{chen2020audionav,chen2021savi}) but, unlike point-goal, the goal location is unknown.
Hence, the agent needs to leverage learned scene priors to explore the environment efficiently to find and navigate to the target.
Current approaches tackle each navigation task separately: a new model is trained for each task and each target modality~\cite{chaplot2020neural,batra2020objectnav,chen2021savi,wu2019}, which has the disadvantages discussed above.
In contrast, we propose a unified approach to semantic visual navigation, in which a \emph{single} trained policy can handle diverse tasks and goal modalities.

\vspace{-0.05cm}
\paragraph{Transfer Learning in Navigation}
Pre-learning a representation from large-scale image datasets~\cite{deng2009imagenet,lin2014coco} and transferring it to a downstream task proved to be very successful for visual recognition~\cite{girshick2014,zamir2018taskonomy,chen2020mocov2,caron2020swav,chen2020simclr}.
We observe a similar trend in embodied navigation, where pre-learning good representations of the 3D environment~\cite{mousavian2019,muller2018driving,zhou2019,yang2018,ramakrishnan2021environment} or primitive skills~\cite{gordon2019splitnet,wijmans2020ddppo,li2020,eysenbach2018} help the agent to learn a downstream task better while using fewer training samples.
Recent methods focus on pre-training the observation encoder of the agent, either in a supervised ~\cite{sax2019,gordon2019splitnet,shen2019situational,landi2021,yen2020learning} or self-supervised~\cite{du2021crl,chen2020mocov2} fashion.
While this leads to improved performance on the target tasks, a new policy is still learned from scratch for each task, resulting in low sample efficiency.
In contrast, our approach enables a full transfer paradigm where all the agent's components can be reused efficiently on the downstream tasks.
Prior work shows that transferring a strong point-goal policy to non-goal driven tasks (\eg, flee and exploration) can lead to better performance ~\cite{wijmans2020ddppo}.
Differently, we propose to learn and transfer a general-purpose semantic search policy.
Our policy can find semantic goals presented in different modalities for a diverse set of goal-driven navigation tasks.

Sharing knowledge between multiple tasks can be achieved in a multi-task learning setup~\cite{wang2020multi,chaplot2019multi} where all tasks are learned jointly in a supervised manner, or via meta-RL~\cite{finn2017maml,wortsman2019meta} where a meta policy learned from a distribution of tasks is finetuned on the target.
Unlike these methods, our policy is learned from one task that does not require manual annotations, and it can be transferred in a zero-shot setup where the policy does not receive any interactive training on the target.

\vspace{-0.1cm}
\paragraph{Zero-Shot Learning}
Zero-shot learning (ZSL) can be seen as an extreme case of transfer learning where the target task has zero training samples.
Prior ZSL work focuses on  supervised learning, \eg, image classification ~\cite{larochelle2008,lampert2009,elhoseiny2013,al-halah2017,rohrbach2011,al-halah2016,xian2017}.
In contrast, the proposed \emph{zero-shot experience learning} (ZSEL) setup learns \emph{behaviors} rather than classifiers; the  policy learned on a source task needs to perform a set of target tasks, without receiving any new interactive experiences on the target.
Further, unlike~\cite{sekar2020} where a world model for synthetic environments is constructed, and control policies are trained on `imagined' episodes, we consider a model-free approach and a ZSEL setup in realistic environments where the policy receives zero target interactions (\ie, neither imagined nor real).
To our knowledge, we are first to propose a ZSEL model for embodied navigation.

\begin{figure*}[t]
\centering
\includegraphics[width=1.\linewidth]{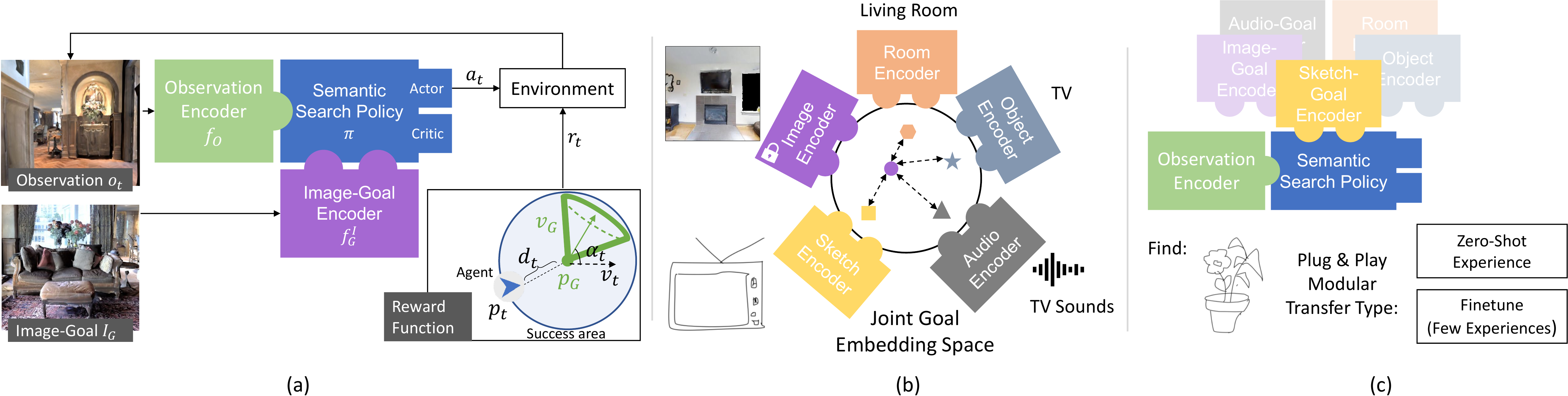}
\vspace*{-0.6cm}
\caption{
    Our approach (a) starts by learning a semantic search policy using a novel reward function for finding random image views in a 3D scene.
    Then, (b) we learn a joint goal embedding space for various goal modalities where the learning is guided by the image-goal encoder.
    Finally, (c) we transfer our model in a plug \& play fashion to a new target task where it can perform out of the box (zero-shot) or it is finetuned using few experiences on the target task. 
}
\label{fig:model}
\vspace*{-0.6cm}
\end{figure*}
 
\section{Plug \& Play Modular Transfer Learning}\label{sec:approach}
We introduce a novel transfer learning approach for visual navigation.
Our model has three main components:
1) we start by learning a semantic search policy for image goals using a novel reward and task augmentation (\figref{fig:model}a);
2) we leverage the image goal encoder to learn a joint goal embedding space for the different goal modalities (\figref{fig:model}b);
and finally, 3) we transfer the learned agent modules to downstream tasks in a plug and play fashion (\figref{fig:model}c).

In the following, we consider an agent with $3$ main modules:
1) an observation encoder ($f_O$) that encodes the received observations $o_t$ from the environment;
2) a goal encoder ($f_G$) that encodes the task's goal;
3) a policy ($\pi$) that uses the output of $f_O$ and $f_G$ to navigate and find the goal.

\subsection{Semantic Search Policy for Image Goals}\label{sec:app_policy}
The policy is a key component in the modern end-to-end visual navigation agent.
It guides the agent towards solving a task given a set of sequential observations and a goal.
Such policies are often learned with reinforcement learning (RL) where the agent interacts with its environment (by moving about) and attempts to solve the task in a trial and error fashion.
If the agent succeeds in its attempt, then it receives a reward to encourage such behavior from the policy in the future.

A main challenge for this learning paradigm is that the policy requires a large number of interactions with the environment in order to find a proper way to solve the task.
This usually amounts to tens and hundreds of millions~\cite{mezghani2021,maksymets2021thda} and up to billions~\cite{wijmans2020ddppo} of interactions, and correspondingly days or weeks of GPU cluster time.
Furthermore, for each new task a policy is typically learned from scratch, which further increases the learning cost substantially.

We propose to learn a general-purpose \emph{semantic search} policy that can be transferred and perform well on a variety of navigation tasks.
Our idea is to learn such a policy with the image-goal task, where the agent receives a picture taken at some unknown camera pose somewhere in the environment, and must travel to find it.
Our choice of image-goal for the source policy is significant.
It requires no manual annotations, and image-goals can be sampled freely anywhere in a training environment.
As a result, the policy can be trained on large-scale experience (\eg, collected from a fleet of robots deployed in various environments) which can improve its generalization to new tasks and domains.
Furthermore, an image-goal encourages the learned policy to capture semantic priors for finding things in a 3D space.
For example, by seeking images of couches and chairs, the agent learns implicitly to leverage these objects' context and the room layout in order to find the image views effectively.

\paragraph{Task Definition}
In an episode of image goal navigation, the agent starts from a random position $p_0$ in an unexplored scene, and it is tasked to find a certain location $p_G$ given an image $I_G$ sampled with the camera at $p_G$.
The agent receives an RGB observation $o_t$ at each step $t$ and needs to perform the best sequence of actions  $a_t \in$ {\small $\{\texttt{move\_forward}, \texttt{turn\_left}, \texttt{turn\_right}, \texttt{stop}\}$} that would bring it to the goal within a maximum number of steps $S$.
Unlike the common point-goal task where the goal location is known~\cite{anderson2018evaluation}, here $p_G$ is unknown, and the agent needs to leverage the learned semantic priors to search and find where $I_G$ could have been sampled from.

Our setup differs from recent methods in ImageNav where  panoramic \ang{360} FoV sensors are required~\cite{mezghani2021,chaplot2020neural,kwon2021}.
Here, we consider a standard \ang{90} FoV for the agent's view~\cite{zhu2017}.
While having a complete FoV sensor simplifies localization, this strong requirement is often not available in common robotic platforms~\cite{locobot,turtlebot,hellorobot} and leads to high computational cost.
This reduces the scalability and adoption of such methods by diverse agent configurations.
In addition, our task setup allows our model to transfer to a diverse set of semantic navigation tasks in a plug and play fashion without the need for modifications to the target tasks (for which the literature does not use panoramic images).

\paragraph{View Reward}
It is common to use the reduced distance to the goal to reward the agent for getting closer to $p_G$ in addition to the success reward of finding and stopping within a small distance $d_s$ of $p_G$.
However, while this reward proves to be quite successful for navigation tasks like point-goal, we argue it is less suited for semantic goals like images.
Since the reward does not carry a signal about the semantic goal itself, the agent may fail or require much more experience in order to capture the implicit relation between the goal and the distance to goal reward (DTG).
For example, if the goal shows an image of an oven, the agent may get close and stop nearby while looking at a book on the counter, and nonetheless receive a full success reward. 
This may lead to capturing trivial or incoherent associations between the goal and the agent's observations.

In order to encourage the agent to leverage the information provided in the goal description $I_G$ and effectively capture useful semantic priors that may help it in finding $p_G$, we propose a new reward function that rewards the agent for looking at $I_G$ when getting closer to $p_G$, so it can better draw the association between its $o_t$ and $I_G$.
Specifically, we define the reward function at step $t$ as:
\begin{equation}\label{eq:reward}
r_t = r_d(d_t , d_{t-1}) + [d_t \leq d_s] \  r_\alpha(\alpha_t , \alpha_{t-1}) - \gamma,
\end{equation}
where $r_d$ is the reduced distance to the goal from the current position relative to the previous one, $r_\alpha$ is the reduced angle in radians to the goal view from the current view relative to the previous one, $[\cdot]$ is the indicator function, and $\gamma=0.01$ is a slack reward to encourage efficiency.
Note, this reward will encourage the agent to look at $I_G$ when it gets near the goal, since it is rewarded to reduce the angle between its current view $v_t$ and the view of the goal $v_G$ (see \figref{fig:model}a).
Finally, the agent receives a maximum success reward of $10$ if it reaches the goal and stops within a distance of $d_s$ from $p_G$ and an angle $\alpha_s$ from $v_G$:
\begin{equation}\label{eq:reward_succ}
R_s = 5 \times ([d_t \leq d_s]  + [d_t \leq d_s \,\mathrm{and}\, \alpha_t \leq \alpha_s]).
\end{equation}
We set $d_s=\SI{1}{\m}$ (the success distance of the task) and $\alpha_s=\ang{25}$ to allow for a good overlap between  $v_t$ and $v_G$ and enable the agent to draw the association between its observation $o_t$ and the goal $I_G$.

\paragraph{View Augmentation}
In addition to the view reward introduced above, we also provide a simple task augmentation method to promote generalization by increasing the diversity of goals presented to the agent.
For each training episode, rather than having a fixed $I_G$, we sample a view from a random angle at location $p_G$ and provide the agent with the associated $I_G$ from the sampled view as the goal descriptor.
This has a regularization effect on the model learning; the agent will be less likely to overfit due to the changing goal description each time the agent experiences a given goal.
Furthermore, with the start $p_0$ and the goal location $p_G$ fixed in a training episode but not $I_G$, this encourages the agent to capture implicit spatial semantic priors of things that usually appear near each other as viewed from $p_G$.
For example, the agent would learn that an image of chairs as seen when peeking from the door is likely to be at the same location of the current image goal that is showing a dining table, since the agent experienced the same episode before but with $I_G$ showing the chairs, hence prompting the agent to explore the dining room.

\paragraph{Policy Training}
We train our policy using reinforcement learning (RL), \figref{fig:model}a.  For each training episode, we sample an image-goal $I_G$ from $p_G$.
The agent encodes its current observation $o_t$ (an RGB image) with $f_O$ and the image-goal with $f_G^I$ and passes these encodings to the policy $\pi$.
The policy further encodes these information along with the history of observations so far to produce a state embedding $s_t$.
An actor-critic network leverages $s_t$ to predict state value $c_t$ and the agent's next action $a_t$.
Based on the agent's state in the environment, it receives a reward (\eqref{eq:reward} and \eqref{eq:reward_succ}). The model is trained end-to-end using PPO~\cite{schulman2017ppo}.

\subsection{Joint Goal Embedding Learning}\label{sec:app_goal}
Having learned the semantic search policy, we can now transfer our model to downstream tasks.  Specifically, we consider downstream navigation tasks where the goals are  \emph{object categories} (ObjectNav~\cite{batra2020objectnav}), \emph{room types} (RoomNav~\cite{wu2019}), or \emph{view encodings} (ViewNav), and they may be expressed by the modalities of a label name, a sketch, an audio clip, or an edgemap; see Sec.~\ref{sec:eval_transfer}.
A key advantage of learning the semantic search policy using RGB image-goals is that these goals contain rich information about the target visual appearance and context.
Furthermore, in order to solve the image-goal navigation task, our model learns to encode these visual cues via a compact dense representation produced by the image-goal encoder $f_G^I$.

Our idea is to leverage $f_G^I$ to learn a joint embedding space of different goal modalities for the various tasks.
In other words, we upgrade the image-goal embedding space to be a joint goal embedding space to draw associations between the images and the different goal modalities like sketches, category names, and audio (\figref{fig:model}b).
This step can be carried out quite efficiently and using an offline dataset.
For example, to learn about object-goals that are represented with a label (\eg, \emph{a chair}) we only need to annotate a set of images with chairs.
Then we train an object-goal encoder to produce an embedding similar to the image-goal encoder for compatible image-label pairs.
In our experiments, we use offline datasets of size $20$K images or less in which the ``annotations" are actually automatic object detections.
This is several orders of magnitude smaller than the amount of interactions usually needed to train a target-specific policy (tens to hundreds of millions)~\cite{maksymets2021thda,ye2021auxiliary}.

Formally, let $D=\{(x_i, g_i)\}$ be a set of images $x_i$ and their associated goals $g_i$, 
where $g_i$ can be of any goal modality (\eg, audio, sketch, image, category name, edgemap) depending on the downstream task specifications.
We learn a joint goal embedding space by minimizing the loss:
\begin{equation}\label{eq:embedding}
\mathcal{L}(x_i, g_i) = \left\{
    \begin{array}{ll}
        \scriptstyle 1 - \mathrm{cos}(f_G^I(x_i), f_G^M(g_i)), & \scriptstyle \mathrm{if} \quad y_i = +1 \\
        \scriptstyle \mathrm{max}(0, \mathrm{cos}(f_G^I(x_i), f_G^M(g_i))), & \scriptstyle \mathrm{if} \quad y_i = -1 \\
    \end{array}\right.
\end{equation}
where $f_G^M(\cdot)$ is the new goal encoder of modality $M$, $\mathrm{cos(\cdot,\cdot)}$ is the cosine similarity between two embeddings, and $y_i$ indicates whether the pair $(x_i, g_i)$ is similar or not, as derived from the offline annotations (\eg, \emph{chair} and a picture of a chair are similar; the audio of TV
and a picture of the TV
are similar).

During goal embedding learning, we freeze $f_G^I$ and learn $f_G^M$ using \eqref{eq:embedding} such that $f_G^M$ learns to encode its goal similar to the corresponding image embedding from $f^I_G$.

\subsection{Transfer and Zero-Shot Experience Learning}\label{sec:app_transfer}

Having learned the semantic search policy and the joint goal embedding as described above, now we can transfer our model to downstream navigation tasks (\figref{fig:model}c).
For that, we only need to replace $f_G^I$ with the suitable goal encoder for the task, such that:
\begin{equation}\label{eq:transfer}
a_t \sim \pi(f_O^I(o_t), f_G^M(g)).
\end{equation}

Our plug and play modular transfer approach has multiple advantages.
Since all modules are compatible with each other, this means the model can perform the target task out of the box, \ie, it does not require any further task-specific interactions to solve the target task.
We refer to this setup as zero-shot experience learning (ZSEL).
Training policies with modern RL frameworks is the most expensive part of the model learning, and with ZSEL we manage to circumvent this requirement.
Furthermore, due to the modular nature of our approach, it is easy to generalize to a wide variety of tasks and goal modalities.
For a new task, only the respective goal encoder is trained with an offline dateset then integrated in the full model in a plug\&play fashion.

Finally, our model can be easily finetuned for the downstream task to capture any additional cues specific to the task to reach a better performance.
Unlike the common approach in the literature where only $f_O$ is pretrained and transferred~\cite{sax2019,zhou2019,du2021crl,gordon2019splitnet}, here the full model is transferred to the target task.
This leads to higher initial performance, faster convergence, and better overall performance, as we will show in~\secref{sec:eval}.

\section{Evaluation}\label{sec:eval}
In the following experiments, we first evaluate our semantic search policy performance in the source task (image-goal navigation) compared to state-of-the-art methods (\secref{sec:eval_imagenav}); then we show how our model transfers to a diverse set of downstream navigation tasks (\secref{sec:eval_transfer}).

\paragraph{Shared Implementation Setup}
For fair comparisons, we adopt the same architecture and training pipeline for our model and all RL baselines, and we note any deviations from this shared setup in the respective sections.
We use a ResNet9~\cite{he2016resnet,shacklett21bps} for $f_O$ and a GRU~\cite{cho2014gru} of $2$ layers and embedding size of $128$ for $\pi$.
For goal encoders $f_G$, we use a ResNet9 to encode image-, sketch-, edgemap- and audio-goal modalities.
We transform an audio clip to a spectrogram before encoding it by $f_G$.
If the goal is a category name, we use a $2$ layer MLP for $f_G$.
We train the policy using DD-PPO~\cite{wijmans2020ddppo} and allocate the same computation resources for all models.
We use input augmentation (random cropping and color-jitter) during training to improve the stability and performance of the RL methods~\cite{laskin2020,kostrikov2021,mezghani2021}.
See Supp for more details.
We adopt end-to-end RNN-based RL since it is a  common~\cite{mezghani2021,wijmans2020ddppo,gordon2019splitnet,du2021crl,zhu2017,zhou2019,sax2019}, generic architecture, does not require hand-crafted modules, shows good performance on real-data~\cite{mirowski2018,chancan2020mvp}, and learned in sim has the potential to generalize well to real~\cite{kadian2020sim2real}. However, our contributions are orthogonal to the RL architecture used.

\paragraph{Agent Configuration}
The action space of the agent consists of {\small\texttt{move\_forward}} by \SI{25}{\cm}, {\small\texttt{turn\_left}} and {\small\texttt{right}} by \ang{30}, and {\small\texttt{stop}}.
The agent uses only RGB observations of $128\times128$ resolution and \ang{90} FoV sensor.

\subsection{Image-Goal Navigation}\label{sec:eval_imagenav}
\vspace{0.2cm}

\begin{table}[t]
\setlength{\tabcolsep}{6pt}
\center
\scalebox{0.85}{
\begin{tabular}{l c cc}
\toprule
    Model                                               & Split &  Succ.      & SPL \\
\midrule                
    Imitation Learning                                  & A     &  9.9        &  9.5  \\
    Zhu \etal~\cite{zhu2017}                           & A     &  19.6       &  14.5 \\
    Mezghani \etal~\cite{mezghani2021} w/ \ang{90} FoV                 & A     &  9.0        &  6.0  \\
    DTG-RL                                              & A     &  22.6       &  18.0 \\
    Ours                                                & A     &  {\bf29.2}  &  {\bf21.6} \\
    Ours (View Aug. Only)                               & A     &  22.0       &  18.8 \\
    Ours (View Reward Only)                             & A     &  24.4       &  17.3 \\
\midrule                
    Hahn \etal~\cite{hahn2021}                          & B     &  24.0       & 12.4 \\
    Ours                                                & B     &  {\bf33.0}  & {\bf23.6} \\
\midrule                
    Hahn \etal~\cite{hahn2021} w/ noisy actuation  & B     &  20.3       & 8.8  \\
    Ours w/ noisy actuation                        & B     &  {\bf25.9}  & {\bf17.6} \\
\bottomrule
\end{tabular}
}
\vspace{-0.2cm}
\caption{
    Image-goal navigation results on Gibson~\cite{xia2018gibson}.
}
\label{tbl:imagenav_main}
\vspace{-0.5cm}
\end{table}
 
\paragraph{Task Setup}
We adopt the image-goal task as defined in~\secref{sec:app_policy}.
We set $S=1000$ and $d_s=\SI{1}{\m}$ from $p_G$.

\paragraph{Datasets}
We use the Habitat simulator~\cite{savva2019habitat} and the Gibson~\cite{xia2018gibson} environments to train our model.
We use the dataset from~\cite{mezghani2021}.
The training split contains $9$K episodes sampled from each of the $72$ training scenes.
Following the setup from~\cite{mezghani2021}, all RL models are trained for $50$K updates ($500$ million frames) on the training split.
The test split has $4.2$K episodes sampled uniformly from $14$ disjoint (unseen) scenes.
For direct comparison with~\cite{hahn2021}, we also test our model on a second split (``split B") provided by~\cite{hahn2021} that has $3$K episodes and the same structure as the test split from~\cite{mezghani2021} (``split A").

\paragraph{Baselines}
We compare our image-goal model to the following baselines and SoTA methods:
1) \textbf{Imitation Learning}: This model's policy is trained using supervised learning to predict the ground truth best action on the shortest path to the goal given its current observation.
2) \textbf{Zhu \etal~\cite{zhu2017}}: The model uses a ResNet50 shared between $f_O$ and $f_G$, pretrained on ImageNet and frozen.
3) \textbf{Mezghani \etal~\cite{mezghani2021}}: This is the SoTA panoramic image-goal navigation model.
It uses a ResNet18 for $f_O$ and $f_G$, a $2$ layer LSTM~\cite{hochreiter1997lstm} for $\pi$, and a specialized episodic memory.
We adapt this model to our \ang{90} FoV for $o_t$ and $I_G$ and train it using the author's code.
4) \textbf{DTG-RL}: This model uses the shared architecture along with the common distance to goal dense reward for training.
5) \textbf{Hahn \etal~\cite{hahn2021}}: This model learns from a passive dataset of videos collected from the Gibson training scenes and uses a customized architecture based on topological maps (see~\cite{hahn2021} for details).

\paragraph{Results and Analysis}
\tblref{tbl:imagenav_main} reports the overall performance in terms of average success rate (Succ) and Success weighted by inverse Path Length (SPL) over $3$ random seeds.
Our model outperforms strong baselines and the SoTA in image-goal navigation by a significant margin.
In split A, our model gains $+6.6\%$ in Succ and $+2.6\%$ in SPL over the best baseline.
The method designed for panoramic sensors~\cite{mezghani2021} tends to underperform in this challenging setting.
We see a drop in Succ from 69\%~\cite{mezghani2021} to 9\% when using \ang{360} and \ang{90} FoV, respectively, since such methods rely heavily on the \ang{360} FoV for accurate localization.
In split B, our model gains $+9\%$ in Succ and $+11.2\%$ in SPL over \cite{hahn2021}.
It is important to note that the model from \cite{hahn2021} uses a much more complete sensor configuration than our method (pose sensor, RGB and Depth sensors of $480\times640$ resolution, and a \ang{120} FoV) and it is trained offline from passive videos sampled from the simulator.
Nonetheless, our model outperforms~\cite{hahn2021} by a large margin, showing that interactive learning of end-to-end RL models still has an advantage over heuristic and passive approaches.

\paragraph{Ablations}
To validate our contributions from \secref{sec:app_policy}, we test our model performance when removing the view reward or the view augmentation.
As shown in \tblref{tbl:imagenav_main}, we see a degradation in performance whenever one of these components are removed, and the largest gain is realized when they work in tandem.
Additionally, we test our model under noisy actuation.
While methods in split A do not provide results under noisy conditions, \cite{hahn2021} does.
Following the setup from \cite{hahn2021}, we use the noise model from~\cite{chaplot2020neural} that simulates actions learned from a Locobot~\cite{locobot}.
Our model shows robustness to noise and maintains its advantage over the baselines (\tblref{tbl:imagenav_main} bottom).

\begin{table}[t]
\setlength{\tabcolsep}{6pt}
\center
\scalebox{.85}{
\begin{tabular}{l cc cc cc cc cc cc}
\toprule
                                         & \multicolumn{2}{c}{MP3D~\cite{Matterport3D}}  & \multicolumn{2}{c}{HM3D~\cite{ramakrishnan2021hm3d}} \\
    Model                                & Succ. & SPL               &   Succ. & SPL  \\
\midrule
    Imitation Learning                   & 5.3  & 5.1 & 2.0 & 1.9 \\
    Zhu \etal~\cite{zhu2017}            & 9.8  & 7.9 & 4.4 & 2.7 \\
    Mezghani \etal~\cite{mezghani2021} w/ \ang{90} FoV  & 6.9  & 3.9 & 3.5 & 1.9 \\
    DTG-RL                               & 11.0 & 9.0 & 5.5 & 3.7 \\
    Hahn \etal~\cite{hahn2021}           & 9.3  & 5.2 & 6.6 & 4.3 \\
    Ours                                 & {\bf14.6} & {\bf10.8} & {\bf9.6} & {\bf6.3} \\
\bottomrule
\end{tabular}
}
\vspace{-0.2cm}
\caption{Image-goal navigation results on MP3D and HM3D in a cross-domain evaluation setup.}
\label{tbl:imagenav_cross_domain}
\vspace{-0.5cm}
\end{table}
 \paragraph{Cross-Domain Generalization}
Next we test the models trained on Gibson on datasets from Matterport3D (MP3D)~\cite{Matterport3D} and HM3D~\cite{ramakrishnan2021hm3d}.
In addition to the visual domain gap between these datasets, MP3D has more complex and larger scenes than Gibson, and HM3D has high diversity in terms of scene types.
This poses a very challenging cross-domain evaluation setting.
The test split from each dataset has in total $3$K episodes sampled uniformly from $100$ and $18$ scenes for HM3D and MP3D, respectively.
\tblref{tbl:imagenav_cross_domain} shows the results.
Overall, we see a drop in performance for all models in this challenging setting, especially for HM3D since there is high diversity in the $100$ val scenes.
Nonetheless, our model outperforms all baselines on both datasets, showing that our contributions lead to better generalization by encouraging the agent to pay closer attention to the semantic information provided by the goal.

\subsection{Transfer to Downstream Tasks}\label{sec:eval_transfer}
\vspace{0.2cm}

\begin{table}[t]
\setlength{\tabcolsep}{2pt}
\center
\scalebox{.8}{
\begin{tabular}{l l |ccc|c|c}
\toprule
                                            &  Source      &   \multicolumn{3}{c}{ObjectNav}      &   \multicolumn{1}{|c}{RoomNav}        &   \multicolumn{1}{|c}{ViewNav} \\
Model                                       &  Task          & \multicolumn{1}{c}{Label}      &   \multicolumn{1}{c}{Sketch}   & \multicolumn{1}{c}{Audio}     &   \multicolumn{1}{|c}{Label}    & \multicolumn{1}{|c}{Edgemap} \\
\midrule
    Task Expert                                 & -         & 8.0      & 6.7        & 6.6       & 8.9      & 0.8   \\
\midrule
    MoCo v2~\cite{chen2020mocov2} (Gib.)      & SSL       & 10.5     & 9.9        & 8.8       & 9.3      & 1.0   \\
    MoCo v2~\cite{chen2020mocov2} (IMN)    & SSL       & 7.8      & 12.7       & 11.5      & 9.7      & 1.3   \\
\midrule
    Visual Priors~\cite{sax2019}                & SL        & 9.3      & 9.9        & 9.1       & 13.1     & 0.6   \\
    Zhou \etal~\cite{zhou2019}                  & SL        & 15.6     & 7.6        & 9.6       & 10.3     & 0.7   \\
\midrule
    CRL~\cite{du2021crl}                        & RL        & 1.9      & 0.5        & 1.0       & 1.2      & 0.0     \\
    SplitNet~\cite{gordon2019splitnet}          & RL        & 9.0      & 6.5        & 8.8       & 7.7      & 0.6   \\ 
    DD-PPO (PN)~\cite{wijmans2020ddppo}   & RL        & 13.9     & 13.6       & 12.9      & 13.9     & 1.7   \\
\midrule
    Ours (ZSEL)                                 & RL        & 11.3     & 11.4       & 4.4       & 11.2     & 5.4   \\
    Ours                                        & RL        & {\bf21.9}  & {\bf22.0}  & {\bf18.0} & {\bf27.9}  & {\bf7.4}   \\
\bottomrule
\end{tabular}
}
\vspace{-0.2cm}
\caption{
    Transfer learning success rate on downstream semantic navigation tasks.
}
\label{tbl:transfer_main}
\vspace{-0.5cm}
\end{table}
 \paragraph{Tasks}
We consider $3$ target tasks and $4$ goal modalities:

\noindent 1) \textbf{ObjectNav}: The agent is asked to find the nearest instance of one of $6$ categories (\emph{bed}, \emph{chair}, \emph{couch}, \emph{potted-plant}, \emph{toilet}, and \emph{tv}) specified by the goal.
We extend the standard ObjectNav specification~\cite{batra2020objectnav} where the goal is given by its label (\eg, find a \emph{chair}) to goals specified by a hand drawn sketches of the category, or audio produced by the object (\eg, sounds from TV).
In the beginning of an episode, the agent gets either a label, a sketch, or a $4$ second audio clip from a random category.
An episode is successful if the agent stops within \SI{1}{\m} of the goal while using less than $S=500$ steps.
For sketches, we use images of the object categories from the Sketch dataset~\cite{eitz2012hdhso}.
For audio clips we sample sounds from the audio dataset in~\cite{chen2021savi} and the audio heard by the agent is scaled by the distance to the goal (\ie, further away goals have fainter sounds).
While for label goals, the category name is the same during training and testing, for audio and sketches, the goal instances used during training are disjoint from those used in testing.
This poses another challenging dimension for the agent to generalize across in addition to the unseen test scenes.

\noindent 2) \textbf{RoomNav}: The agent is tasked with finding the nearest room of $6$ types: \emph{living-room}, \emph{kitchen}, \emph{bedroom}, \emph{office}, \emph{bathroom}, and \emph{dining-room}.
The goal is a label (find an \emph{office}) and the episode is successful if the agent steps inside the room with the maximum episode length $S=500$~\cite{narasimhan2020roomnav}.

\noindent 3) \textbf{ViewNav}: This task is similar to the image-goal navigation task considered above, except a different modality (an edgemap) represents the goal.
This helps  quantify the model performance in target tasks that are more aligned with the source task but with a substantially different goal modality.
In each episode, the agent receives an edgemap from a random view in the scene and needs to find the goal and stops within \SI{1}{\m} to succeed.
We set $S=1000$ to give the agent enough time to succeed in this challenging task.

\vspace{-0.2cm}
\paragraph{Datasets}
For all target tasks, we use $24$ train / $5$ test scenes from the Gibson~\cite{xia2018gibson} tiny set that has semantic annotations~\cite{armeni20193d}.
These scenes are disjoint from those used in \secref{sec:eval_imagenav}.
We train all methods for up to $20$ million steps on the target tasks and report evaluation performance averaged over $3$ random seeds.
See Supp for details.

To train the goal embedding for the modalities in ObjectNav and RoomNav, we sample $14$K images of objects and $20$K of rooms from the training scenes.
We use the object labels generated by a model from~\cite{armeni20193d} to draw the associations between the images and each modality.
For ViewNav, we sample $170$K views from the training scenes and generate their edgemaps using an edge texture model~\cite{zamir2018taskonomy}.
The number of samples for the offline datasest is driven by the available instances of each goal type in the training scenes. While there is a finite number of rooms and objects, we can sample views freely from any location in a scene.

\begin{figure}[t]
\centering
    \includegraphics[width=0.32\linewidth]{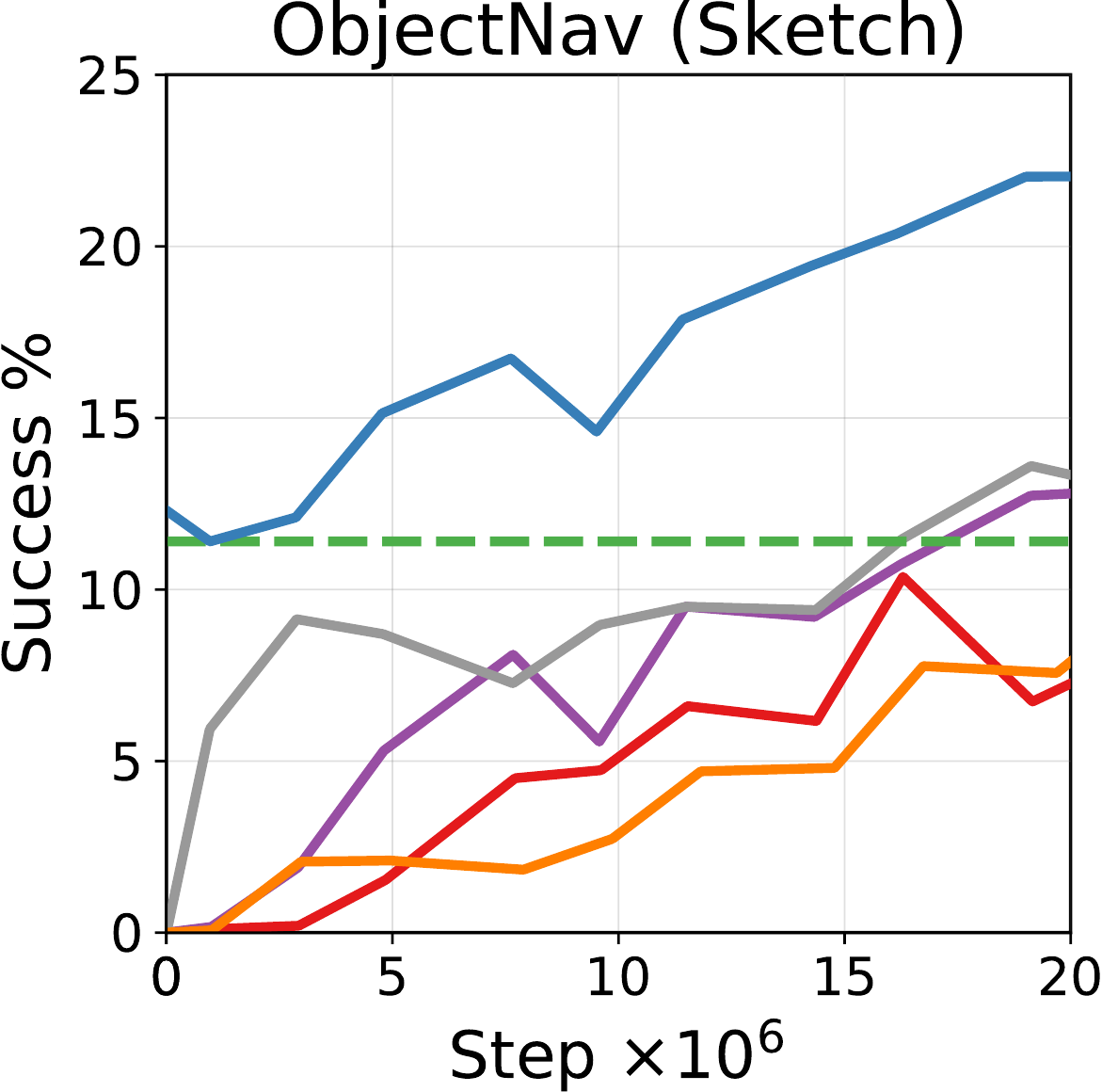}\,
    \includegraphics[width=0.32\linewidth]{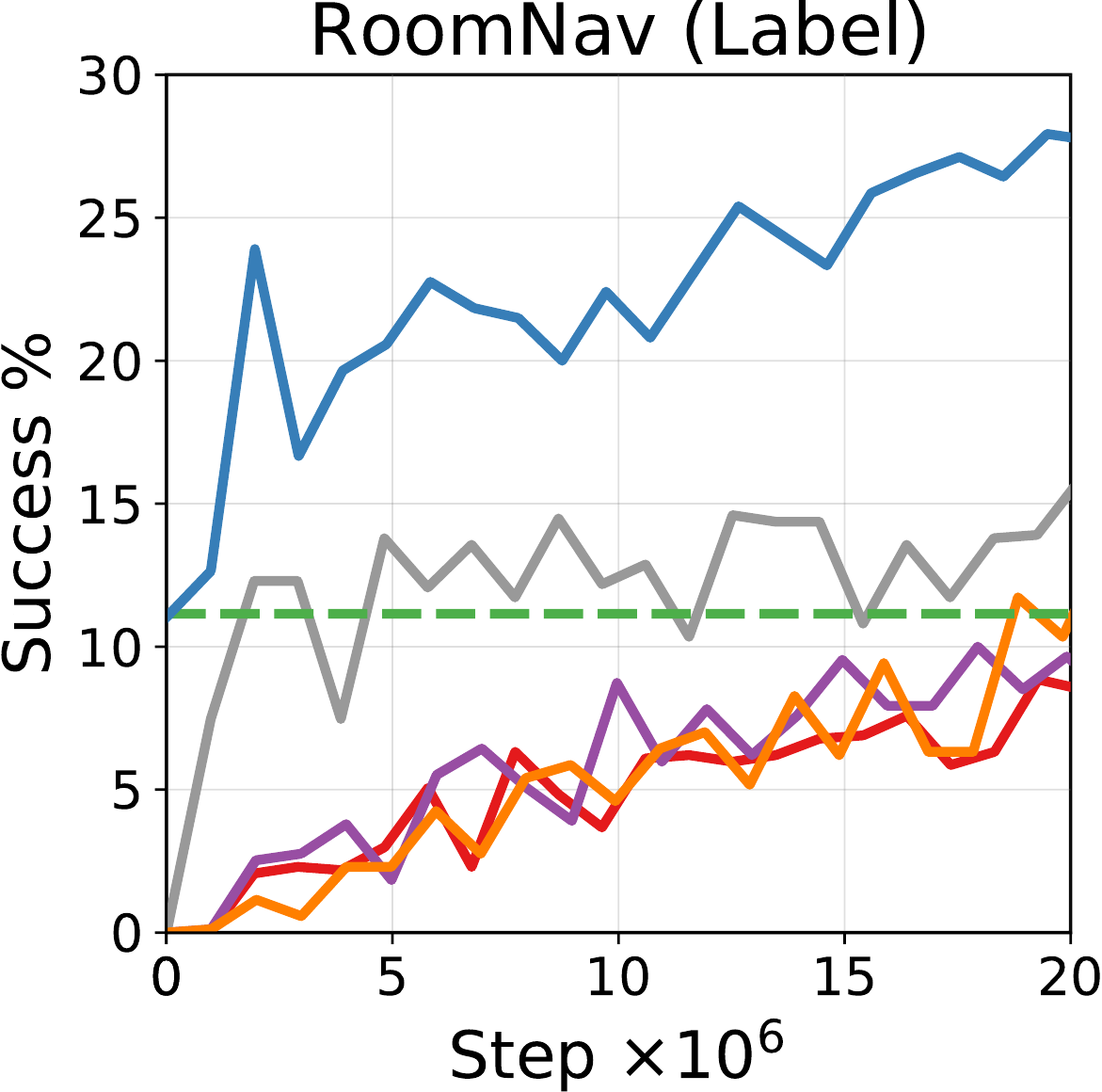}\,
    \includegraphics[width=0.32\linewidth]{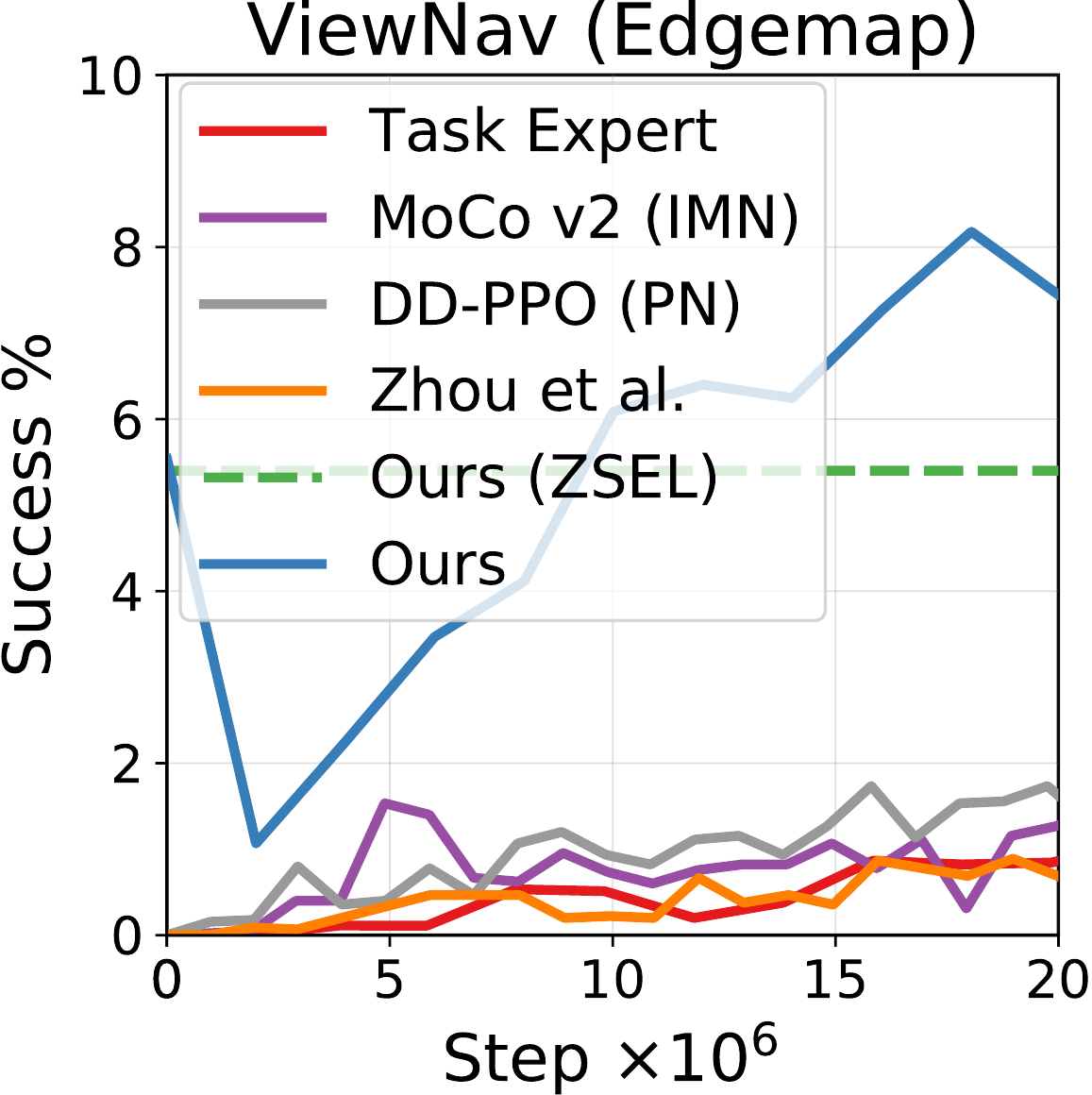}
\caption{
    Transfer learning and ZSEL performance on downstream navigation tasks. See Supp for all tasks and modalities.
}
\label{fig:transfer}
\vspace{-0.5cm}
\end{figure}
 
\paragraph{Baselines}
We compare our model to a set of baselines and SoTA models in transfer learning:
1) \textbf{Task Expert} which learns from scratch on the downstream task.
2) \textbf{MoCo v2}~\cite{chen2020mocov2} initializes $f_O$ using MoCo training on ImageNet (IMN) or on a set of images randomly sampled from the Gibson (Gib.) training scenes.
3) \textbf{CRL}~\cite{du2021crl} pretrains $f_O$ (ResNet50) using a combination of curiosity-based exploration and self-supervised learning.
We initialize $f_O$ from a pretrained model provided by the authors.
4) \textbf{Visual Priors}~\cite{sax2019} uses a set of $4$ ResNet50s pretrained encoders as $f_O$.
The encoders are trained in a supervised manner to predict $4$ features (\eg semantic segmentation, surface normal) that provide maximum coverage for downstream navigation tasks~\cite{sax2019}.
5) \textbf{Zhou \etal}~\cite{zhou2019} transfers $2$ pretrained ResNet50s for depth prediction and semantic segmentation; however, unlike~\cite{sax2019}, these are used with an RGB encoder (ResNet9) trained from scratch.
6) \textbf{SplitNet}~\cite{gordon2019splitnet} pretrains $f_O$ (customized CNN) using a mix of $6$ auxiliary tasks (motion and visual tasks) and point-goal navigation.
We initialize $f_O$ from a pretrained model provided by the authors.
7) \textbf{DD-PPO (PN)}~\cite{wijmans2020ddppo} pretrains the model for point-goal navigation (PN) and both $f_O$ and $\pi$ are transferred.

\paragraph{Transfer Learning}
\tblref{tbl:transfer_main} shows the results.  Our approach outperforms all baselines by a significant margin.
Interestingly,  the self-supervised methods~\cite{chen2020mocov2} reach a competitive performance to those that rely on the availability of dense annotations (like semantic segmentation and ground truth depth) for supervised representation learning~\cite{sax2019,zhou2019}.
Furthermore, methods that learn a curiosity-based representation (CRL~\cite{du2021crl}) or via auxiliary tasks and RL (SplitNet~\cite{gordon2019splitnet}) do not transfer as well as the SSL and SL methods.
Additionally, compared to the strong DD-PPO~\cite{wijmans2020ddppo} approach which was trained on the same data as our policy but for the PointNav task, our model achieves substantial gains in success rate (from $+5\%$ and up to $+14\%$) across all tasks. 
This indicates that our semantic search policy is much better suited to transfer to diverse downstream tasks compared to the PointNav policy.  
Moreover, when looking at the test performance over the course of training for the best transfer methods compared to ours (\figref{fig:transfer}), we notice that our approach has a much higher start and improves faster to better performance.
Our model reaches the top performance of the best competitor up to $12.5\times$ faster.

\begin{figure}[t]
\centering
    \includegraphics[width=0.41\linewidth]{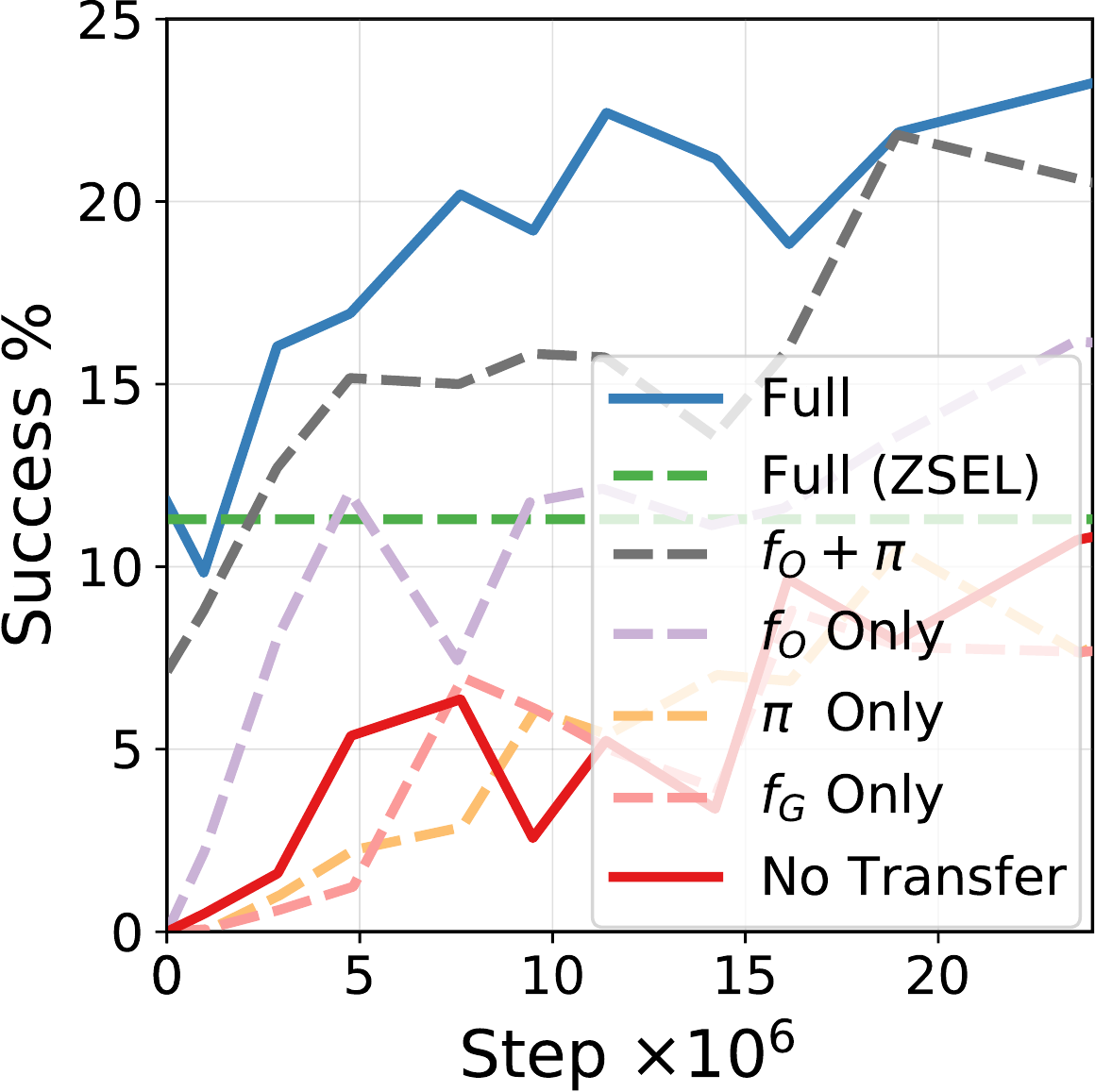}\qquad
    \includegraphics[width=0.33\linewidth]{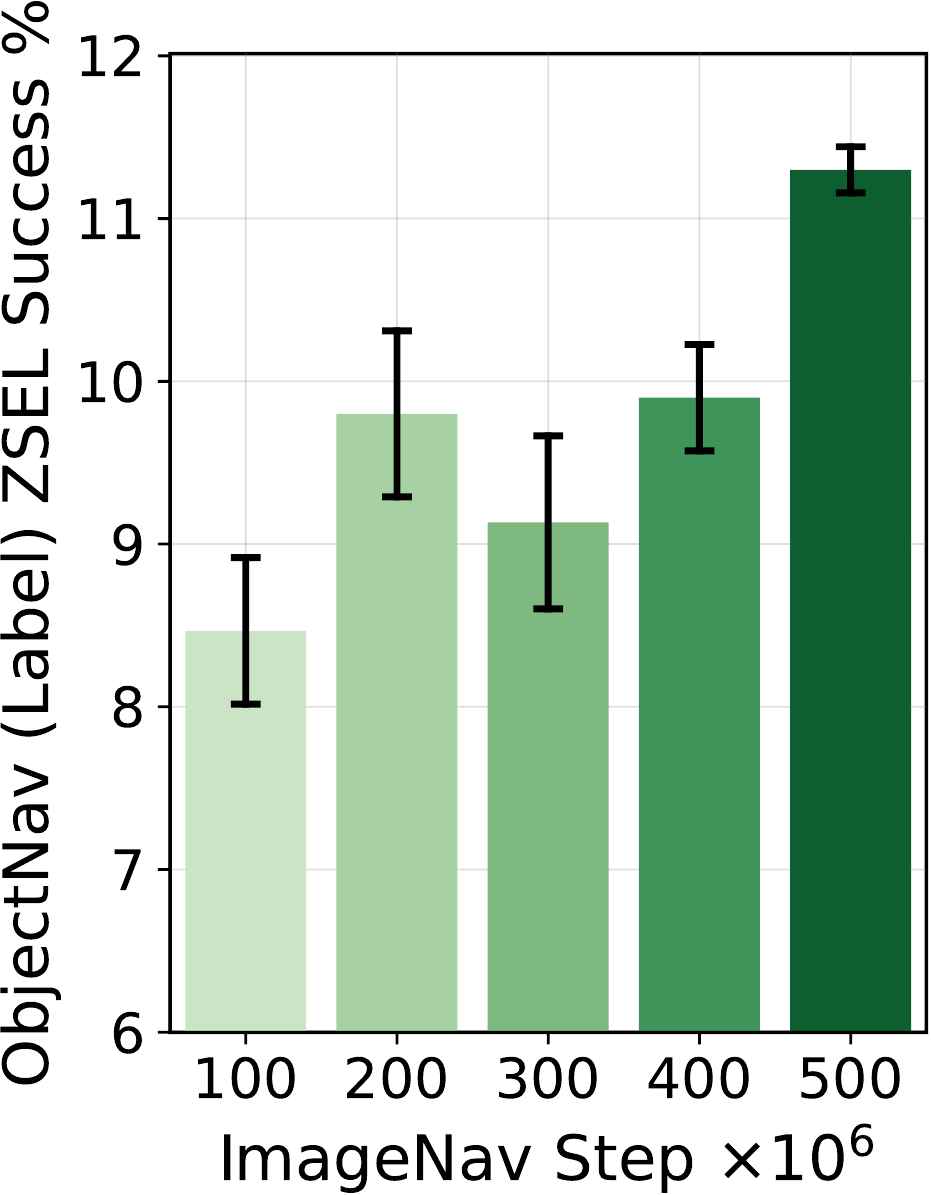}
\vspace{-0.2cm}
\caption{
    Our model ablation for modular transfer (left) and  scalability (right). 
}
\label{fig:ablation}
\vspace{-0.5cm}
\end{figure}

\vspace{-0.2cm}
\paragraph{Zero-Shot Experience Learning}
A unique feature of our approach is its ability to perform the downstream task without receiving any new experiences from it.
Our model shows excellent performance under the challenging ZSEL setting.
Our ZSEL model outperforms the Task Expert in $4$ out $5$ of the tasks despite receiving zero new experience on the target, and even after training the Task Expert for up to $20$ million steps (Ours-ZSEL in \tblref{tbl:transfer_main}).

In addition, we see in \figref{fig:transfer} that the majority of transfer learning models struggle to reach our ZSEL performance.
Note that our model does not have any advantages in terms of architecture, which is shared with the rest of the models.  Thus, the high ZSEL performance is attributable to our modular transfer approach.
In ObjectNav and RoomNav, the best competitor requires between $2$ to $16$ million steps to reach our ZSEL performance, and with the exception of ObjectNav-Audio the competitors show little improvement over that level.
In ViewNav, we notice that none of the baselines are capable of reaching our ZSEL level.
This can be attributed to the challenging goal modality where estimating distances for successful stopping is difficult, and to the close proximity of this task to the source ImageNav task that our semantic policy is most familiar with.

\vspace{-0.1cm}
\paragraph{Modular Transfer Ablation}
\figref{fig:ablation} (left) shows a modular ablation of our approach on the ObjectNav-Label task.
Transferring individual modules separately has mixed impact on performance.
While transferring $f_G$ and $\pi$ only does not improve over the `No Transfer' case, $f_O$ leads to positive transfer effect.
This is expected since in this model $f_O$ is a deep CNN with the largest portion of parameters.
Having a good initialization of this component is beneficial.
Nonetheless, when combining the modules together with our plug and play modular approach we see substantial gains.
Our full model demonstrates the best performance and enables ZSEL, thus validating our contributions.

\begin{figure}[]
\centering
    \includegraphics[width=0.32\linewidth]{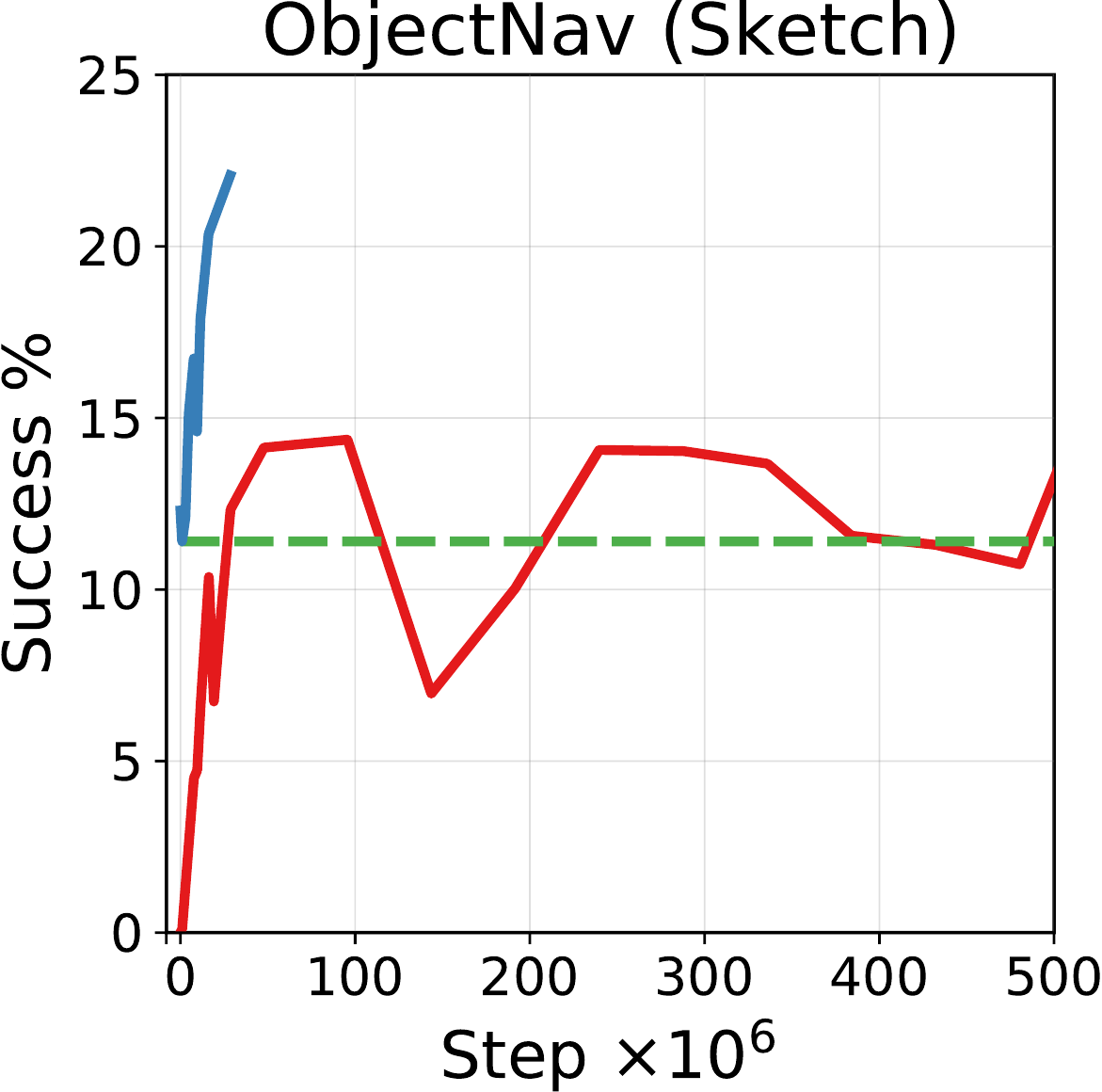}
    \includegraphics[width=0.32\linewidth]{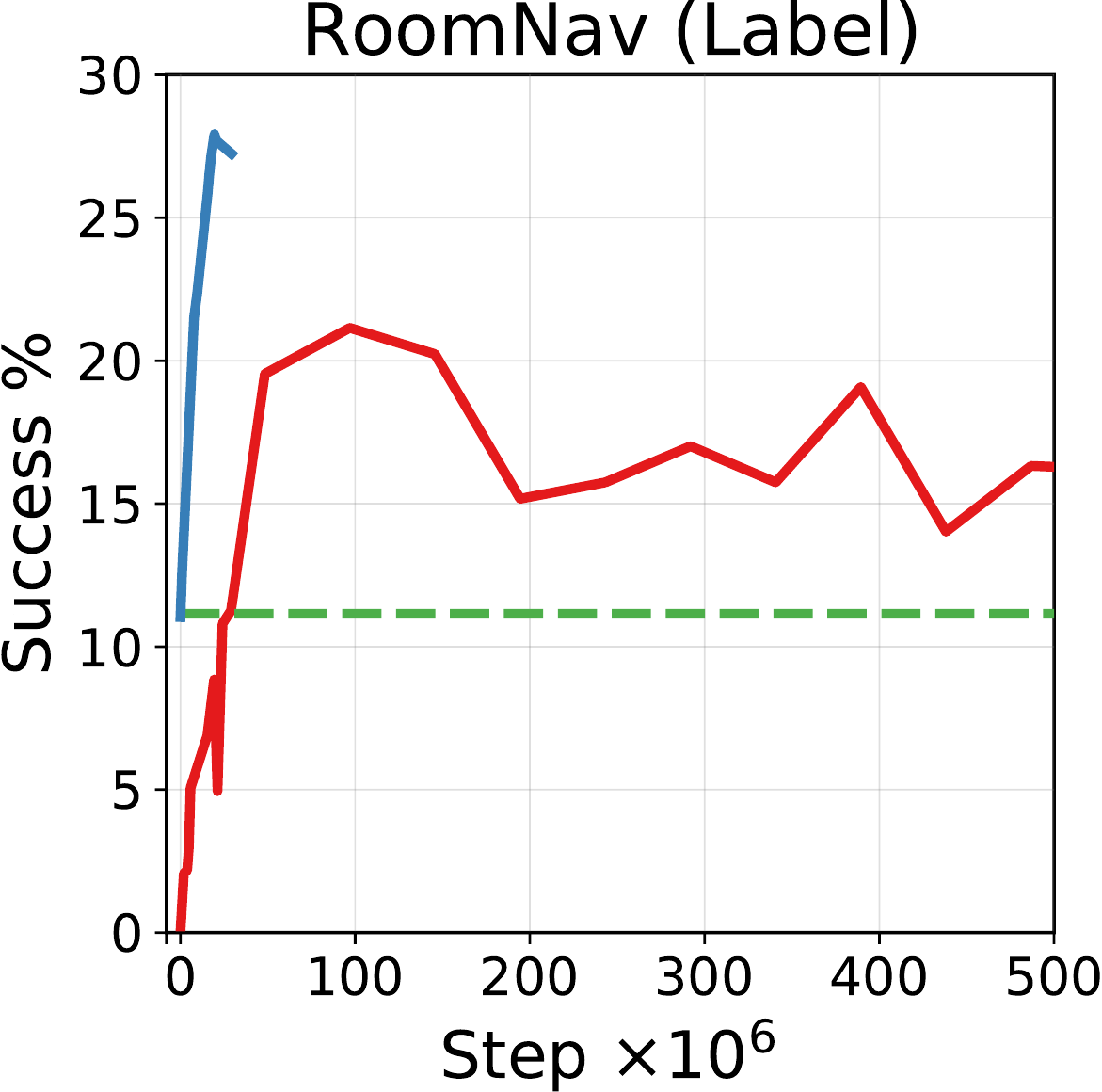}
    \includegraphics[width=0.32\linewidth]{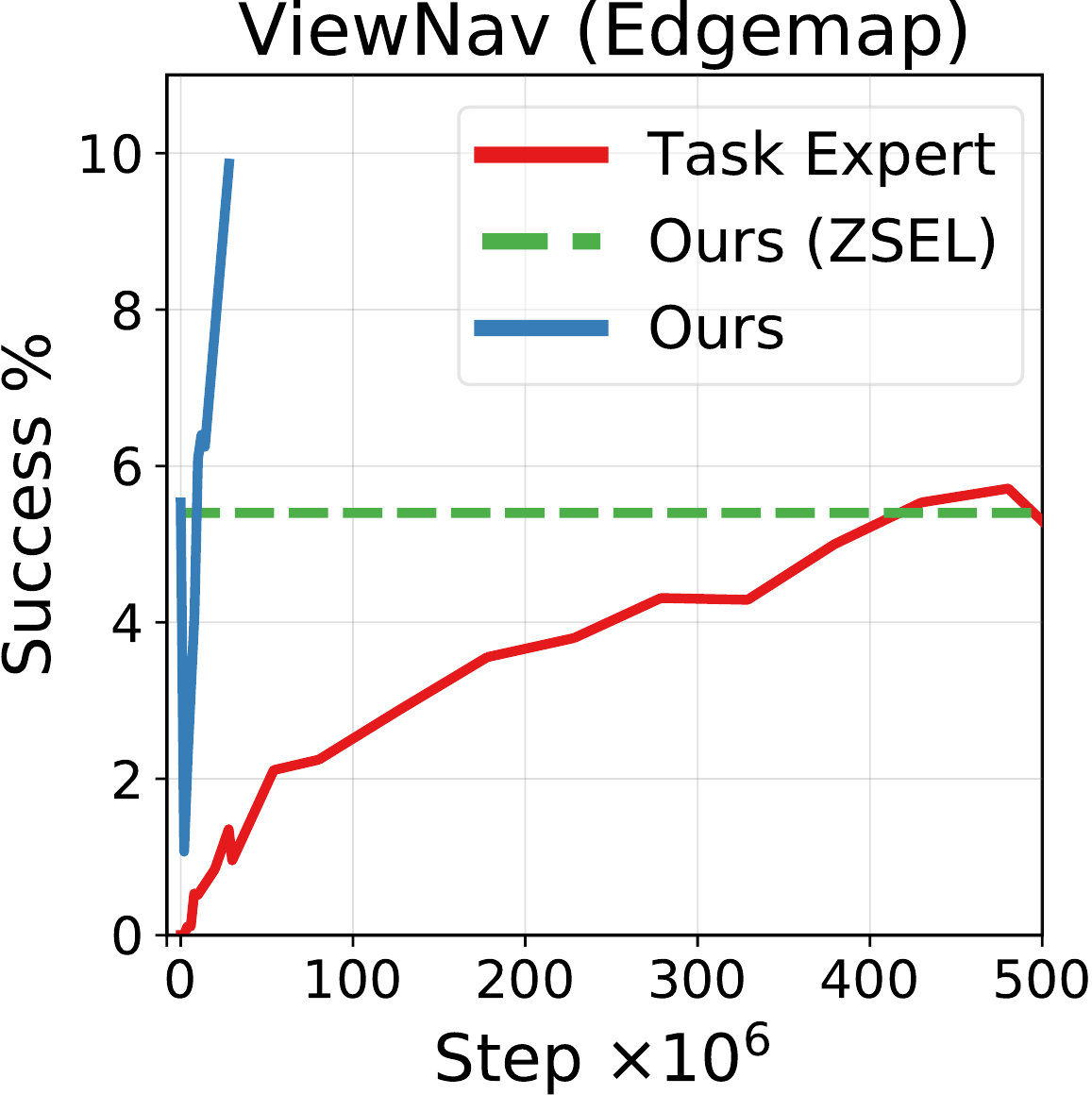}
\vspace{-0.1cm}
\caption{
    Long-term Task Expert training. See Supp for all tasks.
}
\label{fig:longterm}
\vspace{-0.5cm}
\end{figure}
 \vspace{-0.1cm}
\paragraph{Scalability}
We evaluate our model's ability to scale in terms of experience gathered on the source task.
We find a strong correlation between the experience gathered on the source task and the ZSEL performance on downstream ones.
As our semantic search policy receives more experience on the source task (ImageNav), its ZSEL performance on the target task gets better (\figref{fig:ablation} right).
This is important since our source task requires no annotations and can be easily scaled to more scenes and large datasets.
For an analysis of our model in terms of the used sensors, see Supp.

\vspace{-0.1cm}
\paragraph{Long-Term Task Expert Training}
We saw above that our model scales well and its transfer improves when more experience is gathered on the source task.
However, does a task expert become competitive if it simply gets longer training on the target task? How long does that model take to catch up with our approach?
To find out, we train the Task Expert on each of the target tasks for up to $500$M steps.
\figref{fig:longterm} shows the results.
The Task Expert requires on average more than $22$M steps on ObjectNav and RoomNav, and up to $416$M on ViewNav (in total $507$M steps over the $5$ tasks) to reach our ZSEL performance.
It never reaches our model's top performance when our model is finetuned on the target task.
Moreover, our model reaches the best performance of the Task Expert $34.7\times$ faster.
The Task Expert needs task-specific experience with task-specific annotations, which can be expensive and limits the available training data.
In contrast, our model learns in the source task using more diverse goals that can be sampled randomly from the (unannotated) scene, 
thus scaling more effectively.

\vspace{-0.1cm}
\paragraph{Additional Results and Discussions}
Please see Supp for qualitative results, an analysis of failure cases, and a discussion of limitations and the societal impact of our approach.

\section{Conclusion}
\vspace{-0.1cm}
We introduce a plug\&play modular transfer learning approach that provides a unified model for a diverse set of semantic visual navigation tasks with different goal modalities.
Our semantic search policy outperforms the SoTA in the source task of image-goal navigation, as well as the SoTA in transfer learning for visual navigation by a significant margin.
Furthermore, our model is able to perform new tasks effectively with zero-shot experience---to our knowledge, a completely new functionality for visual navigation.
This is a stepping stone for future work, especially for tasks with high-cost training data.
Being able to do ZSEL and learn from few experiences is a crucial skill for an agent in open-world and lifelong learning settings.

\noindent\textbf{Acknowledgements:} UT Austin is supported in part by DARPA L2M, the UT Austin IFML NSF AI Institute, and the FRL Cog Sci Consortium. K.G is paid as a Research Scientist by Meta AI.
Thanks to Lina Mezghani for providing  access to data and code.

{\small
\bibliographystyle{ieee_fullname}
\bibliography{mybib}
}
\vfill
\clearpage
\section{Supplementary Materials}

\begin{figure}[t]
\centering
    \includegraphics[width=0.95\linewidth]{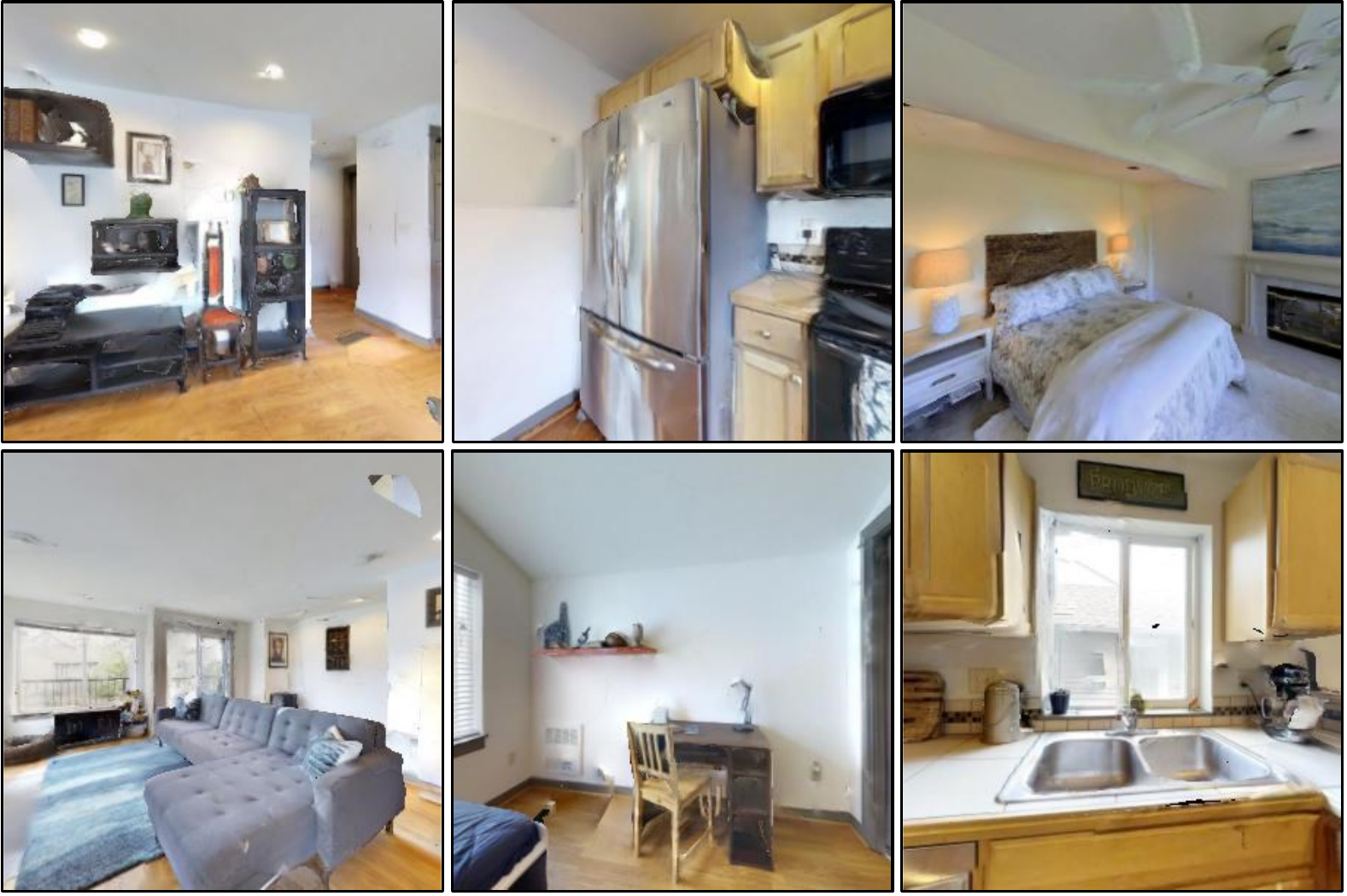}\\\vspace{0.2cm}
    \includegraphics[width=0.95\linewidth]{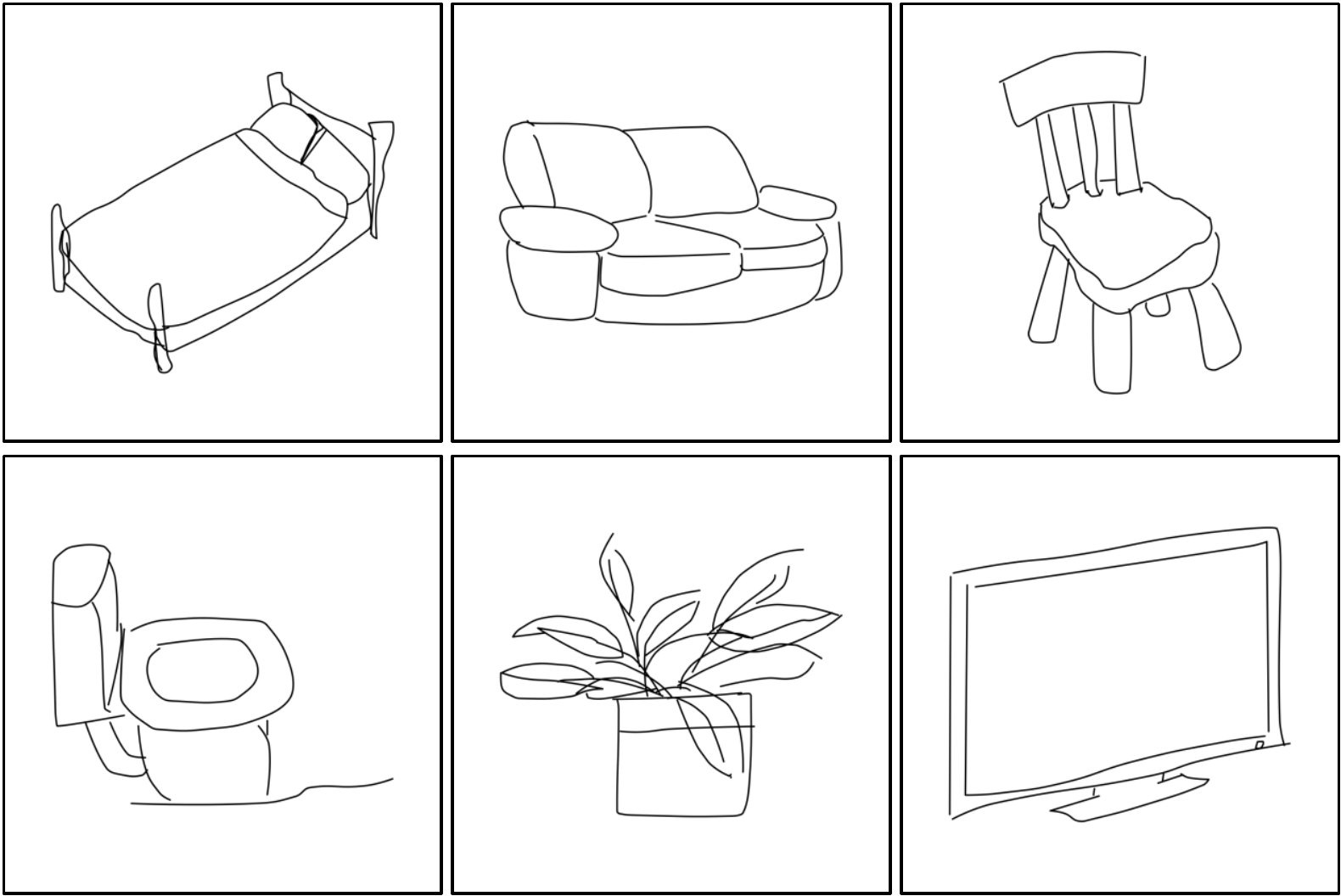}\\\vspace{0.2cm}
    \includegraphics[width=0.95\linewidth]{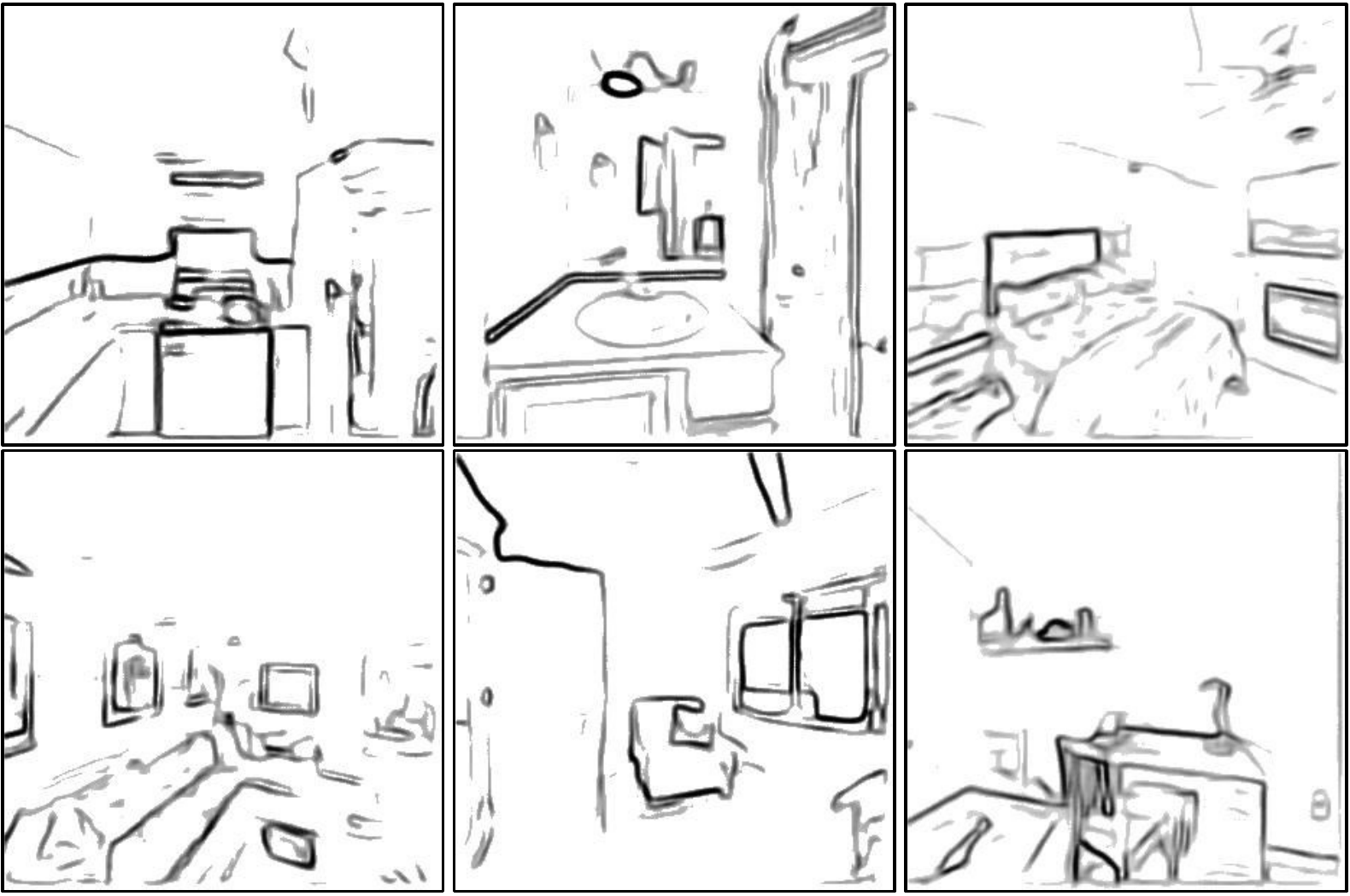}
\caption{
    Examples of visual goals used for the target navigation tasks: top) ImageNav (Image), middle) ObjectNav (Sketch), bottom) ViewNav (Edgemap).
}
\label{fig:goal_examples}
\end{figure}

Additional information presented in this supplementary:
\begin{enumerate}[itemsep=0pt, leftmargin=*]
\item Details on the shared implementation (\secref{sec:shared_setup_supp}).
\item Details of the image-goal navigation dataset (\secref{sec:image_goal_supp}).
\item Detailed results on image-goal navigation across $3$ levels of episode difficulties (\tblref{tbl:imagenav_main}).
\item Examples of the visual goal modalities used in target tasks (\figref{fig:goal_examples}).
\item Dataset details for the target tasks and goal embedding space (\secref{sec:transfer_supp}).
\item Detailed results with standard deviations on target tasks (\tblref{tbl:transfer_main_all}).
\item Qualitative results for our model in all tasks and goal modalities (\figref{fig:qual_all_tasks}).
\item Examples of failure cases for our model (\figref{fig:qual_failure}).
\item Performance curves for all tasks and goal modalities in transfer learning setup (\figref{fig:transfer_all}) and long-term training of Task Expert (\figref{fig:longterm_all}).
\item Ablation on the sensor configuration used by the agent (\figref{fig:scalability_sensors})
\item Discussion of potential societal impact (\secref{sec:impact_supp}) and limitations (\secref{sec:discussion_supp}).
\end{enumerate}

\begin{table*}[h!]
\setlength{\tabcolsep}{7pt}
\center
\scalebox{0.9}{
\begin{tabular}{l c cc cc cc cc}
\toprule
                                    & &   \multicolumn{2}{c}{Easy}        &   \multicolumn{2}{c}{Medium}      &   \multicolumn{2}{c}{Hard}        &   \multicolumn{2}{c}{Overall} \\
    Model                           & Split &   Succ. & SPL &   Succ. & SPL &   Succ. & SPL &   Succ. & SPL \\
\midrule
    Imitation Learning  & A &  18.5 & 17.7 &   8.4  & 8.1  &   2.6  &  2.6  &   9.9  &  9.5  \\
    Zhu \etal~\cite{zhu2017}            & A &  31.7 & 25.1 &   15.7 & 10.8 &   11.5 &  7.5  &   19.6 &  14.5 \\
    Mezghani \etal~\cite{mezghani2021} w/ \ang{90} FoV  & A &  17.5 & 11.0 &   8.8  & 6.6  &   0.6  &  0.5  &   9.0  &  6.0  \\
    DTG-RL                  & A &  32.9 & 26.2 &   21.2 & 17.0 &   13.6 &  10.8 &   22.6 &  18.0 \\
    Ours                                 & A &  {\bf39.7} & 28.5 &   {\bf29.6} & {\bf22.5} &   {\bf18.2} &  {\bf13.8} &   {\bf29.2} &  {\bf21.6} \\
    Ours (View Aug. Only)                & A &  37.0 & {\bf31.7} &   18.5      & 15.9      &   10.7      &  9.0       &   22.0      &  18.8 \\
    Ours (View Reward Only)              & A &  32.3 & 22.7      &   24.8      & 18.1      &   16.1      &  11.0      &   24.4      &  17.3 \\
\midrule
    Hahn \etal~\cite{hahn2021}           & B &  35.5      & 18.4      &   23.9      & 12.1      &   12.5      & 6.8       &   24.0      & 12.4 \\
    Ours                                 & B &  {\bf48.0} & {\bf34.2} &   {\bf36.0} & {\bf25.9} &   {\bf15.1} & {\bf10.8} &   {\bf33.0} & {\bf23.6} \\
\midrule
    Hahn \etal~\cite{hahn2021} w/ noisy actuation  & B &  27.3      & 10.6      &  23.1       & 10.4      &  {\bf10.5}  & {5.6}        &  20.3       & 8.8  \\
    Ours w/ noisy actuation                        & B &  {\bf41.0} & {\bf28.2} &  {\bf27.3}  & {\bf18.6} &   {9.3}     & {\bf6.0}     &  {\bf25.9}  & {\bf17.6} \\
\bottomrule
\end{tabular}
}
\caption{
    Detailed results across $3$ levels of difficulties on image-goal navigation in Gibson~\cite{xia2018gibson}.
}
\label{tbl:imagenav_main}

\end{table*}
 
\begin{table*}[t]
\setlength{\tabcolsep}{8pt}
\center
\scalebox{.9}{
\begin{tabular}{l l | r r r | r | r}
\toprule
                                            &        &   \multicolumn{3}{c|}{ObjectNav}      &   \multicolumn{1}{c|}{RoomNav}        &   \multicolumn{1}{c}{ViewNav} \\
Model                                       &  Source Task          & \multicolumn{1}{c}{Label}      &   \multicolumn{1}{c}{Sketch}   & \multicolumn{1}{c|}{Audio}     &   \multicolumn{1}{c|}{Label}    & \multicolumn{1}{c}{Edgemap} \\
\midrule
    Task Expert                                 & -         & 8.0\std{0.6}      & 6.7\std{1.4}        & 6.6\std{0.7}       & 8.9\std{0.8}      & 0.8\std{0.3}   \\
\midrule
    MoCo v2~\cite{chen2020mocov2} (Gib.)      & SSL       & 10.5\std{0.7}     & 9.9\std{0.6}        & 8.8\std{1.2}       & 9.3\std{0.9}      & 1.0\std{0.2}   \\
    MoCo v2~\cite{chen2020mocov2} (IMN)    & SSL       & 7.8\std{0.3}      & 12.7\std{0.8}       & 11.5\std{0.8}      & 9.7\std{2.2}      & 1.3\std{0.3}   \\
\midrule
    Visual Priors~\cite{sax2019}                & SL        & 9.3\std{0.1}      & 9.9\std{0.7}        & 9.1\std{0.8}       & 13.1\std{0.9}     & 0.6\std{0.1}   \\
    Zhou \etal~\cite{zhou2019}                  & SL        & 15.6\std{1.0}     & 7.6\std{0.3}        & 9.6\std{0.8}       & 10.3\std{0.9}     & 0.7\std{0.1}   \\
\midrule
    CRL~\cite{du2021crl}                        & RL        & 1.9\std{0.5}      & 0.5\std{0.3}        & 1.0\std{0.4}       & 1.2\std{0.8}      & 0.0\std{0.0}     \\
    SplitNet~\cite{gordon2019splitnet}          & RL        & 9.0\std{1.0}      & 6.5\std{0.8}        & 8.8\std{1.1}       & 7.7\std{1.1}      & 0.6\std{0.0}   \\ 
    DD-PPO (PN)~\cite{wijmans2020ddppo}   & RL        & 13.9\std{0.6}     & 13.6\std{0.8}       & 12.9\std{0.5}      & 13.9\std{1.6}     & 1.7\std{0.1}   \\
\midrule
    Ours (ZSEL)                                 & RL        & 11.3\std{0.2}     & 11.4\std{0.6}       & 4.4\std{0.6}       & 11.2\std{1.3}     & 5.4\std{0.6}   \\
    Ours                                        & RL        & {\bf21.9}\std{0.1}  & {\bf22.0}\std{0.9}  & {\bf18.0}\std{1.2} & {\bf27.9}\std{1.9}  & {\bf7.4}\std{0.1}   \\
\bottomrule
\end{tabular}
}
\caption{
    Transfer learning average success rate and standard deviation on downstream semantic navigation tasks.
}
\label{tbl:transfer_main_all}

\end{table*}
 \section{Shared Setup}\label{sec:shared_setup_supp}

All RL methods are trained with the following setup.
We use input augmentation of random cropping and color jitter for both observations and goals.
The models are trained with DD-PPO~\cite{wijmans2020ddppo}.
We set the number of PPO epochs $2$, the forward steps $128$, the entropy coefficient $0.01$, clipping of $0.2$, and train the model end-to-end using the Adam optimizer~\cite{kingma2014adam}.
We allocate the same number of processes and resources to all methods.

We use the Habitat simulator~\cite{savva2019habitat} along with the Gibson~\cite{xia2018gibson}, Matterport3D~\cite{Matterport3D}, and HM3D~\cite{ramakrishnan2021hm3d} datasets.
These datasets are photorealistic and scans of real-world environments with varying complexities, sizes, room layouts, types.
In all our experiments, the test scenes are disjoint from those used for training to assess the agent ability to generalize to previously unseen environments.

\section{Image-Goal Navigation}\label{sec:image_goal_supp}

\paragraph{Dataset}
The training split contains $9$K episodes sampled from each of the $72$ Gibson training scenes.
The episodes are uniformly split across $3$ levels of difficulty based on the goal's geodesic distance from the start location: \emph{easy} (1.5 - \SI{3}{\m}), \emph{medium} (3 - \SI{5}{\m}), and \emph{hard} (5 - \SI{10}{\m}).
Test split A has $4.2$K episodes and split B has $3$K episodes.
Both splits are sampled uniformly from $14$ disjoint (unseen) scenes and the $3$ levels of difficulty.
Further, to avoid trivial straight line paths, the episodes has a minimum geodesic to euclidean distance ratio of $1.1$ in A and $1.2$ in B (our split B corresponds to the curved split in~\cite{hahn2021}).

\paragraph{Detailed Results}
We show in \tblref{tbl:imagenav_main} the detailed results of all models across the three levels of episode difficulties (\emph{easy}, \emph{medium} and \emph{hard}).
Our model shows better performance across the different levels and in both split A and B.
For a qualitative result, see \figref{fig:qual_all_tasks} A.

\section{Transfer Learning to Downstream Tasks}\label{sec:transfer_supp}

\paragraph{Datasets} We use $29$ scenes from Gibson and split them into $24$ scenes for training and $5$ for testing.
In the following, we present the details of the datasets used for each of the downstream tasks.
\begin{itemize}[leftmargin=*]
    \item ObjectNav: We sample $24$K episodes for training and $1$K episodes for testing.
    For the sketch-goals (\figref{fig:goal_examples} middle), we sample $80$ sketches from~\cite{eitz2012hdhso} for each object category and split them to $70$ used during training and $10$ for testing.
    For the audio-goals, we sample $12$ audio clips from~\cite{chen2021savi} of lengths ranging from $13$ to $53$ seconds and split them $50/50$ for training and testing.
    At the start of each episode of the ObjectNav (Audio) task a random $4$ seconds duration is sampled from the respective audio clip and split, and presented to the agent as the goal descriptor.
    \item RoomNav: We sample $25$K episodes for training and $290$ for testing.
    \item ViewNav: We sample $24$K episodes for training and $1.5$K for testing. We generate edgemaps for $300$ random views per scene, and we randomly assign one of those per episode as the goal (\figref{fig:goal_examples} bottom).
\end{itemize}

\begin{figure*}[t]
\centering
    \includegraphics[width=0.196\linewidth]{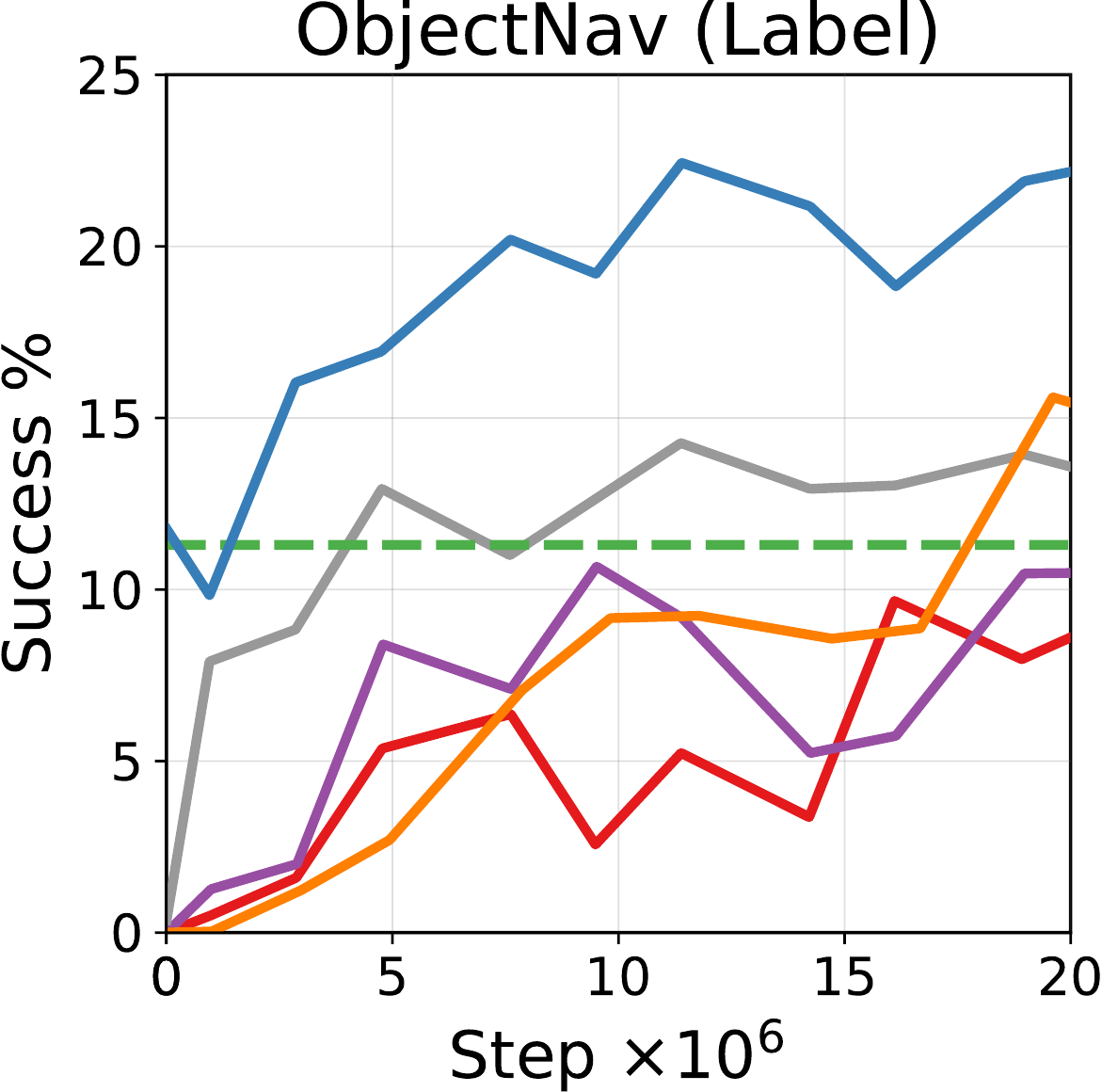}
    \includegraphics[width=0.196\linewidth]{./objectnav_sketch.pdf}
    \includegraphics[width=0.196\linewidth]{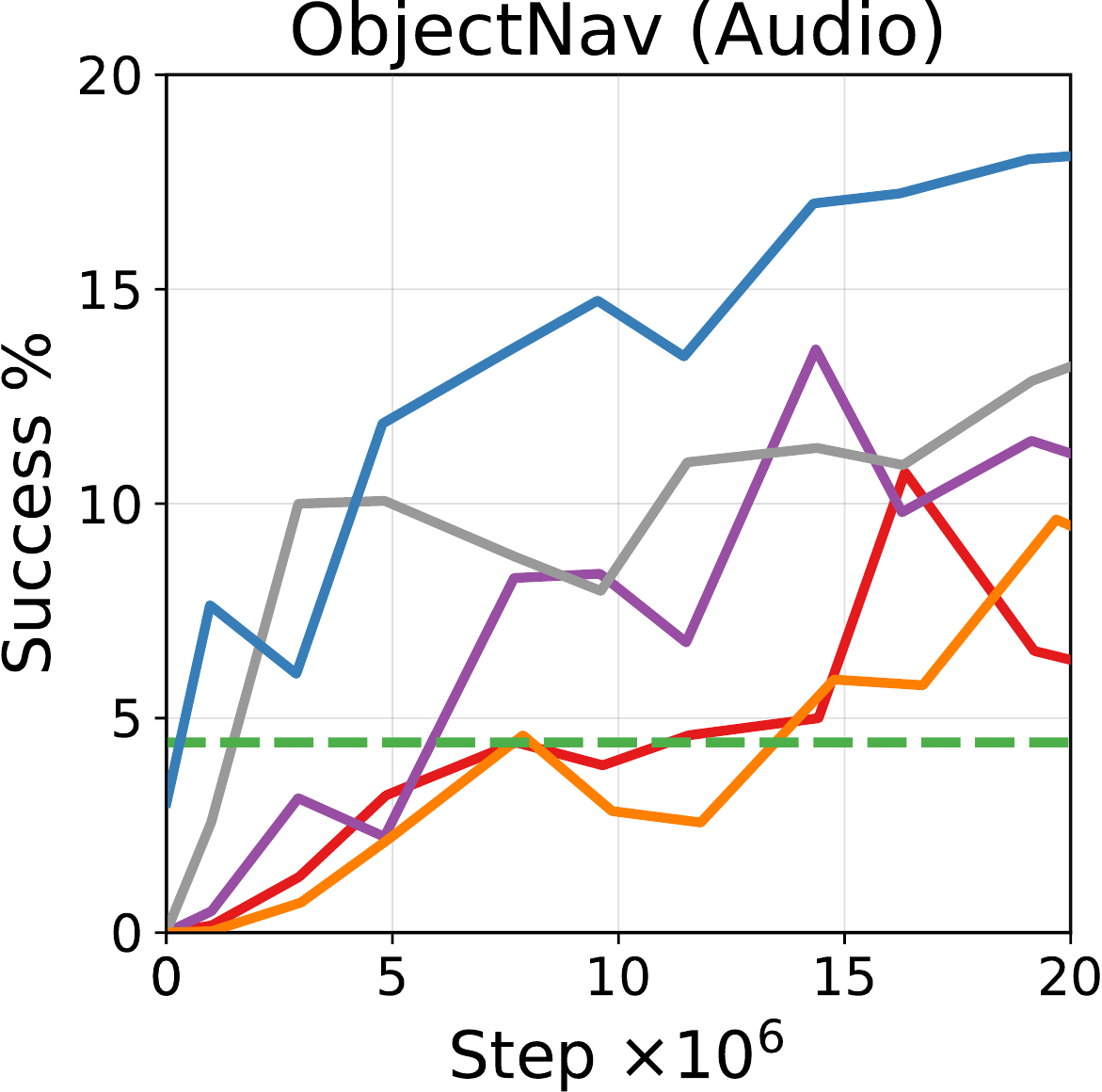}
    \includegraphics[width=0.196\linewidth]{./roomnav_label.pdf}
    \includegraphics[width=0.196\linewidth]{./viewnav_edgemap.pdf}
\caption{
    Transfer learning and ZSEL performance on downstream navigation tasks.
}
\label{fig:transfer_all}
\end{figure*}
 
\begin{figure*}[]
\centering
    \includegraphics[width=0.196\linewidth]{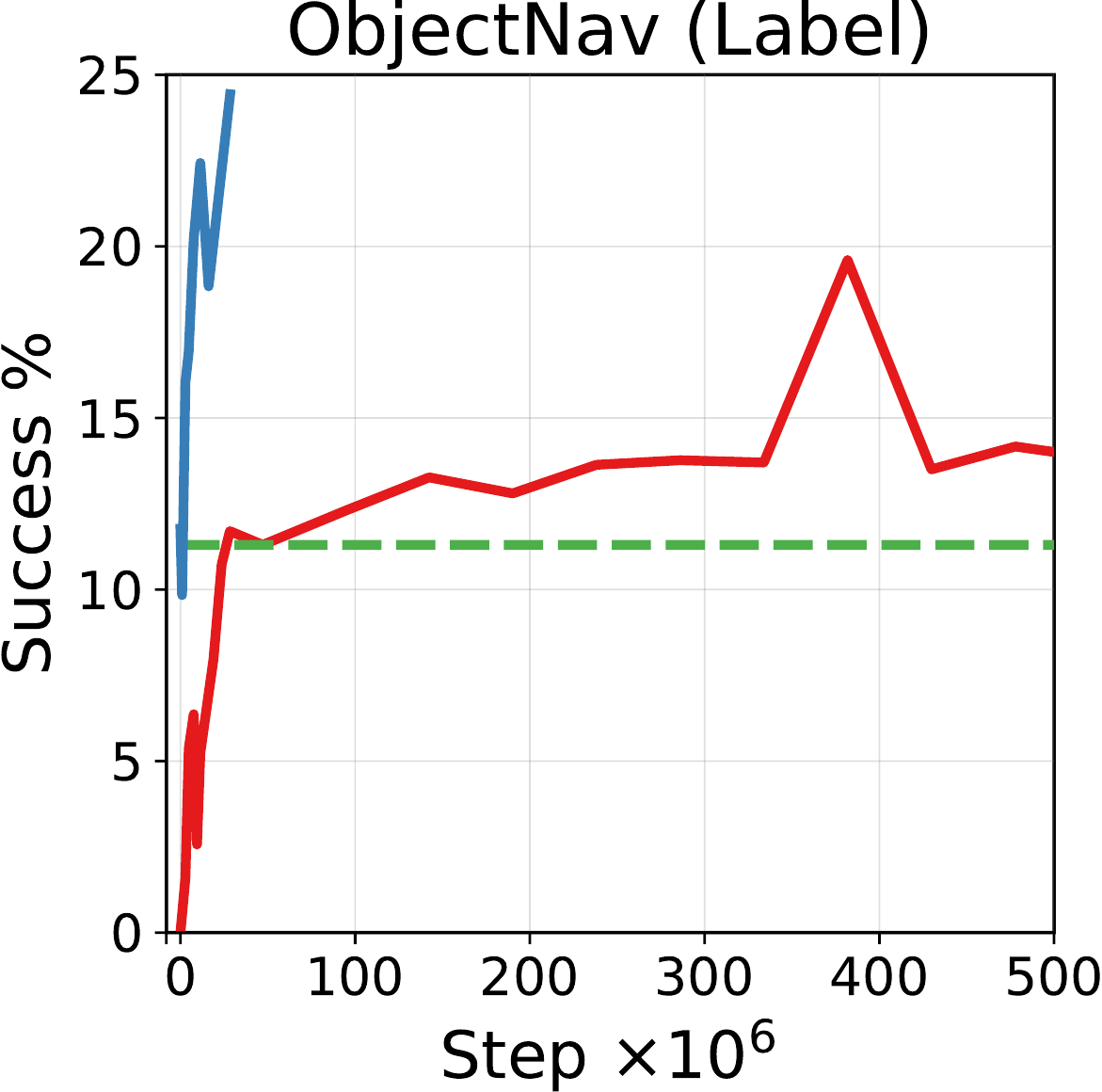}
    \includegraphics[width=0.196\linewidth]{./objectnav_sketch_lt.pdf}
    \includegraphics[width=0.196\linewidth]{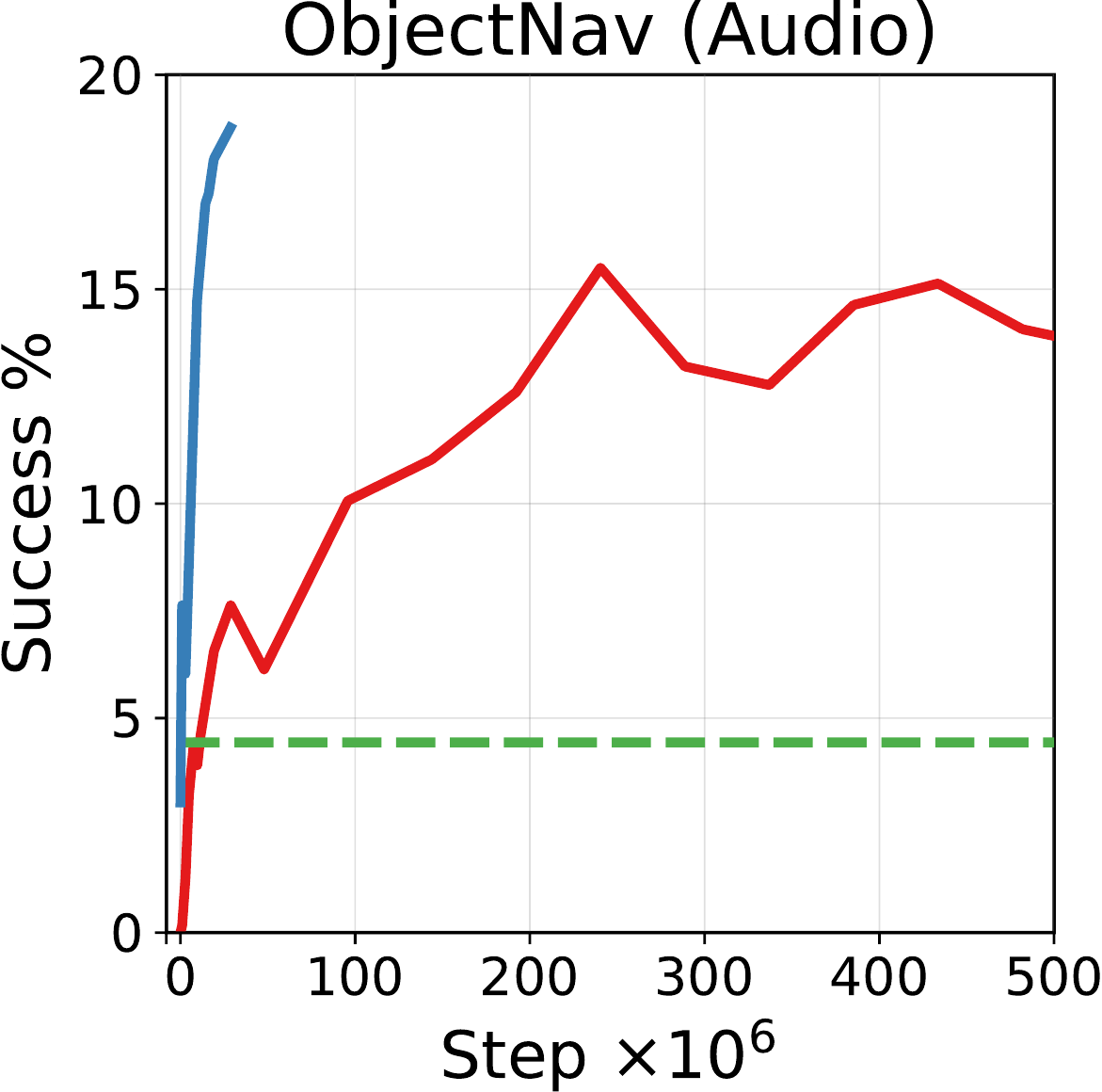}
    \includegraphics[width=0.196\linewidth]{./roomnav_label_lt.pdf}
    \includegraphics[width=0.196\linewidth]{./viewnav_edgemap_lt.pdf}
\caption{
    Long-term Task Expert training.
    Our model maintains it superior performance even when the Task Expert is presented with extensive experience in the target task.
}
\label{fig:longterm_all}
\end{figure*}
 \paragraph{Goal Embedding Space}
In the joint goal embedding space, we aim to learn goal encoders that are compatible to the image-goal encoder.
For example, an image view from a living room with a TV detected in it will be used as the positive anchor for a sketch of a TV, a sound clip from a TV, the TV label, the living room label, and the edgemap of the view.
The annotations for the sampled image view are based on model predictions from~\cite{armeni20193d}.
During training, the parameters of $f_G^I$ are kept frozen, and we train the various goal encoders defined in Main/Sec.4 using the loss from Main/Sec.3.2.

\begin{figure*}[h!]
\centering
    \includegraphics[width=1.0\linewidth]{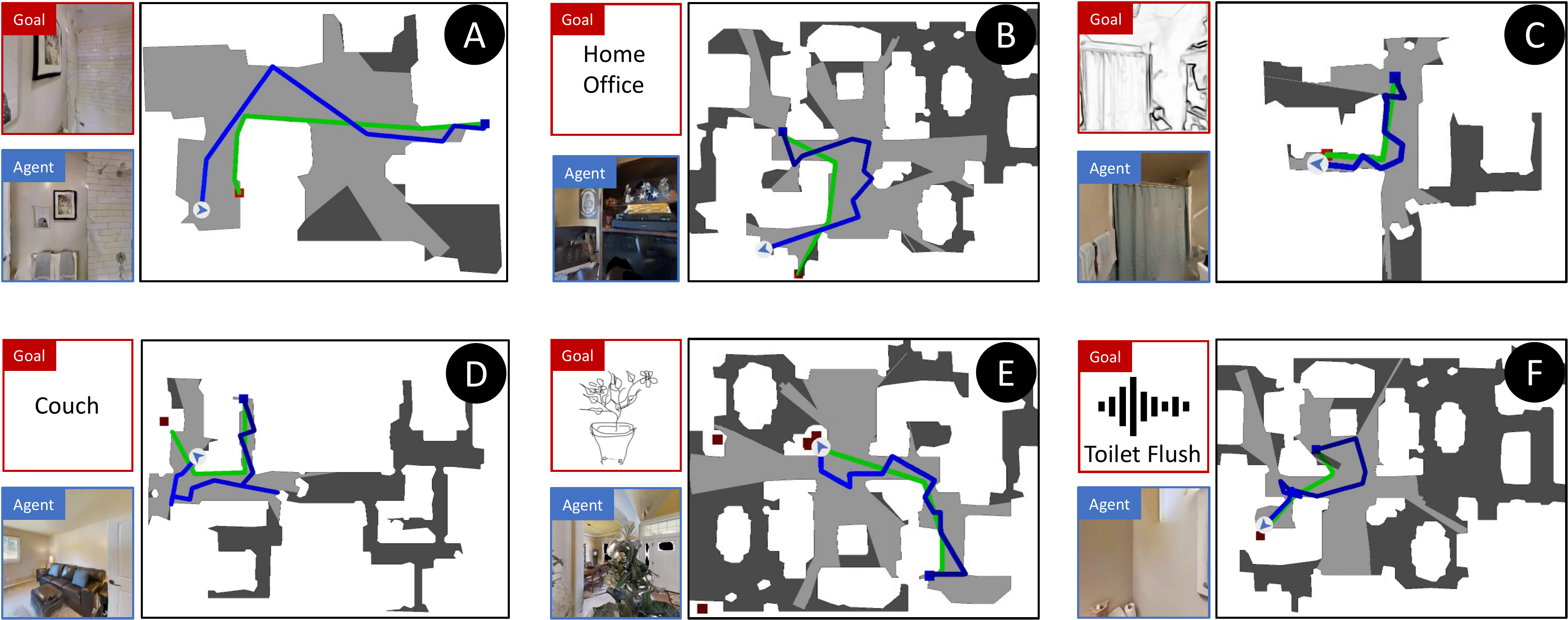}
\caption{
    Qualitative results of our approach performing $6$ tasks with $5$ goal modalities: A) ImageNav (Image), B) RoomNav (Label), C) ViewNav (Edgemap), D) ObjectNav (Label), E) ObjectNav (Sketch), F) ObjectNav (Audio).
}
\label{fig:qual_all_tasks}
\end{figure*}

\paragraph{Detailed Results for All Tasks}
In~\tblref{tbl:transfer_main_all} we show the average success rate and standard deviation for all methods over $3$ random seeds.
\figref{fig:transfer_all} shows the performance of the best transfer learning methods and our approach across all tasks and goal modalities.
\figref{fig:longterm_all} shows the Task Expert performance when trained for up to $500$M steps on each of the respective tasks and in comparison to our model performance under the ZSEL setting or when it is finetuned.
Furthermore, we show example navigation episodes from all tasks and goal modalities for our approach in \figref{fig:qual_all_tasks}.
Our plug and play modular transfer learning approach enable our model to perform a diverse set of tasks effectively.

\begin{figure}[t]
\centering
    \includegraphics[width=0.45\linewidth]{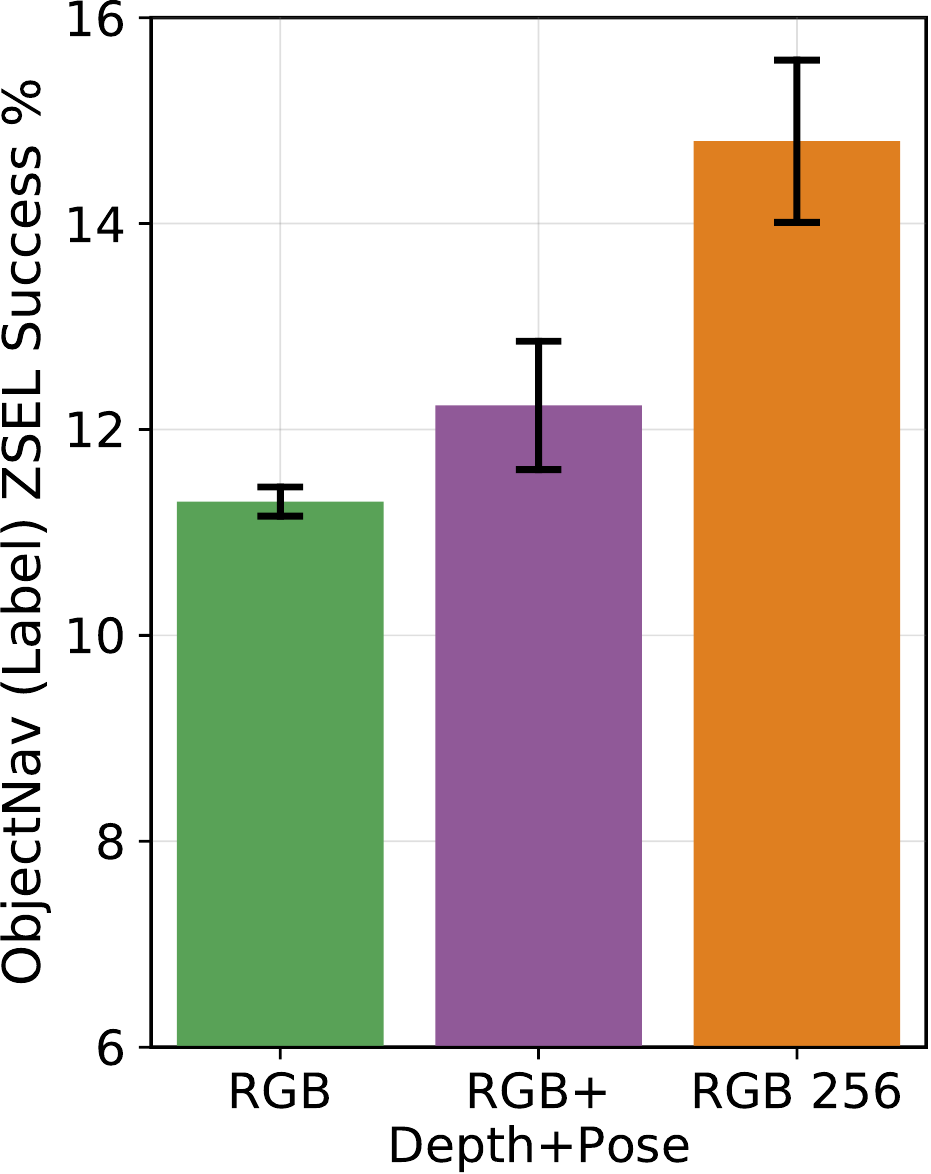}
\caption{
    Scalability ablation for our model when changing the sensors configuration.
}
\label{fig:scalability_sensors}
\end{figure}
 
\begin{figure*}[h!]
\centering
    \includegraphics[width=1.0\linewidth]{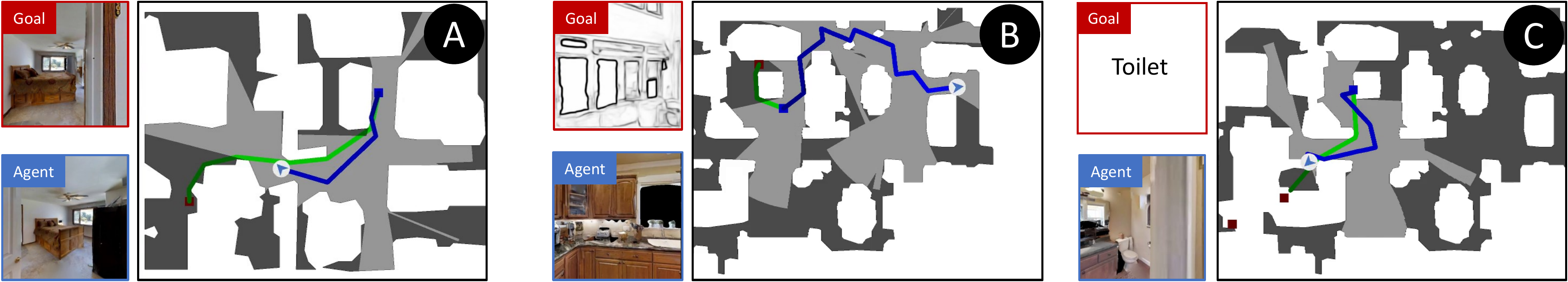}
\caption{
    Qualitative results of failure cases in A) ImageNav (Image), B) ViewNav (Edgemap), and C) ObjectNav (Label).
}
\label{fig:qual_failure}
\end{figure*}
 \paragraph{Scalability (Sensors)}
We evaluate our model's ability to scale across the sensor suite.
\figref{fig:scalability_sensors} shows our model performance when varying the sensors' configuration in the source task (ImageNav) and evaluating on ObjectNav (Label) under the ZSEL setup.
As expected, when enriching the agent sensors to include depth and pose sensors in addition to vision, we see an additional improvement in performance.
More importantly, when increasing the vision sensor resolution from $128$ to $256$ we see a significant bump in ZSEL success rate that exceeds the one from diversifying the sensory suite.
Our model seems to benefit from an enhanced vision channel as it carries the important semantic cues needed for our semantic search policy and goal embedding space.

\paragraph{Failure Cases}
We show in \figref{fig:qual_failure} few examples of failure cases encountered by our model.
We notice that some of these failure cases are related to the type of the goal modality.
For example, in ImageNav the agent sometimes finds the object described in the image however misestimate the view point the image is taken from, hence stops a bit far from the goal location (\figref{fig:qual_failure} A).
In ViewNav (Edgemap), the goal modality lacks distinctive texture and color information which leads the agent to sometime stops at a location with similar edge structure, but it is actually not the goal (\figref{fig:qual_failure} B).
A type of failure cases spotted in multiple tasks are the early stopping cases.
In these cases, the agent fails to estimate the distance to the goal correctly and stops early resulting in an unsuccessful episode (\figref{fig:qual_failure} C).

\section{Potential Societal Impact}\label{sec:impact_supp}
Our approach's application domain is semantic visual navigation.
Here, autonomous agents are trained to find semantic objects in a 3D environment.
Such a technology can have positive societal impact by improving people's life, especially in domains like elder care, with robots that can aid in daily life tasks (\eg find my keys, go to the bedroom and bring me my medicine).
On the other side, the datasets used in this study are 3D scans of building and houses from certain geographic and cultural areas (western style houses from well-off areas).
This creates certain biases in the type of building architectures, room, and object types the agent is familiar with.
Consequently, this may limit the availability of this technology to a small section of the population.
More diverse datasets and methods with robust adaptation to strong shifts in building layouts and object types are needed to mitigate these effects.

\section{Discussion and Limitations}\label{sec:discussion_supp}
We propose a novel approach for modular transfer learning that enables the agent to handle multiple tasks with diverse goal modalities effectively.
Our model can solve the downstream tasks out-of-the-box in zero-shot experience learning setup alleviating the need for expensive interactive training of the policy.
Alternatively, our model can be finetuned on the downstream task to learn task-specific cues where it showed to learn faster, generalize better and reach higher performance than the baselines.
While we focused in this work on semantic navigation tasks, this can be seen as a first step in this exciting direction.
Additional research is needed to generalize this method to tasks that require a series of goals and a compatible policy that can plan effectively in a multi-goal setup (\eg VLN~\cite{anderson2018vln}).
Further, our results and evaluation demonstrate strong transfer learning performance for our method. However, as usual in transfer learning, there is not a theoretical guarantee that a transfer effect will always be beneficial.
Target tasks with significant differences to the source task may not benefit from transferring the accumulated experience in the source.
 
\end{document}